\documentclass[preprint,12pt,authoryear]{elsarticle}

\usepackage{amssymb}
\usepackage{graphicx}
\usepackage{dcolumn}
\usepackage{bm}
\usepackage{mathrsfs}
\usepackage{soul}
\usepackage{scalerel}
\usepackage{tikz}
\usepackage{subfig}
\usepackage{caption}

\usepackage{url}

\journal{Machine Learning with Applications}

\begin{document}

\begin{frontmatter}



\title{Multi-module based CVAE to predict HVCM faults in the SNS accelerator}
\author[1]{Yasir Alanazi\corref{cor1}}
\ead{alanazi@jlab.org}
\author[1]{Malachi Schram}
\ead{schram@jlab.org}
\author[1]{Kishansingh Rajput}
%
\author[1]{Steven Goldenberg}
%
\author[1]{Lasitha Vidyaratne}
%
%
%
\author[2]{Chris Pappas}
%
\author[3]{Majdi I. Radaideh}
%
\author[2]{Dan Lu}
%
\author[2]{Pradeep Ramuhalli}
%
\author[2]{Sarah Cousineau}
%
\cortext[cor1]{Corresponding author}
\affiliation[1]{organization={Thomas Jefferson National Accelerator Facility},
city={Newport News}, postcode={Virginia 23606}, country={USA}}
\affiliation[2]{organization={Oak Ridge National Laboratory},
city={Oak Ridge}, postcode={Tennessee 37830}, country={USA}}
\affiliation[3]{organization={Department of Nuclear Engineering and Radiological Sciences, The University of Michigan}, city={Ann Arbor},
postcode={Michigan 48109}, country={USA}}
\begin{abstract}
We present a multi-module framework based on Conditional Variational Autoencoder (CVAE) to detect anomalies in the power signals coming from multiple High Voltage Converter Modulators (HVCMs).
We condition the model with  the specific modulator type to capture different representations of the $normal$ waveforms and to improve the sensitivity of the model to identify a specific type of fault when we have limited samples for a given module type. 
We studied several neural network (NN) architectures for our CVAE model and evaluated the model performance by looking at their loss landscape for stability and generalization. 
Our results for the Spallation Neutron Source (SNS) experimental data show that the trained model generalizes well to detecting multiple fault types for several HVCM module types.
The results of this study can be used to improve the HVCM reliability and overall SNS uptime. 
\end{abstract}




\end{frontmatter}


\section{Introduction}  
\begin{figure}
\includegraphics[width=\textwidth, height = 7cm]{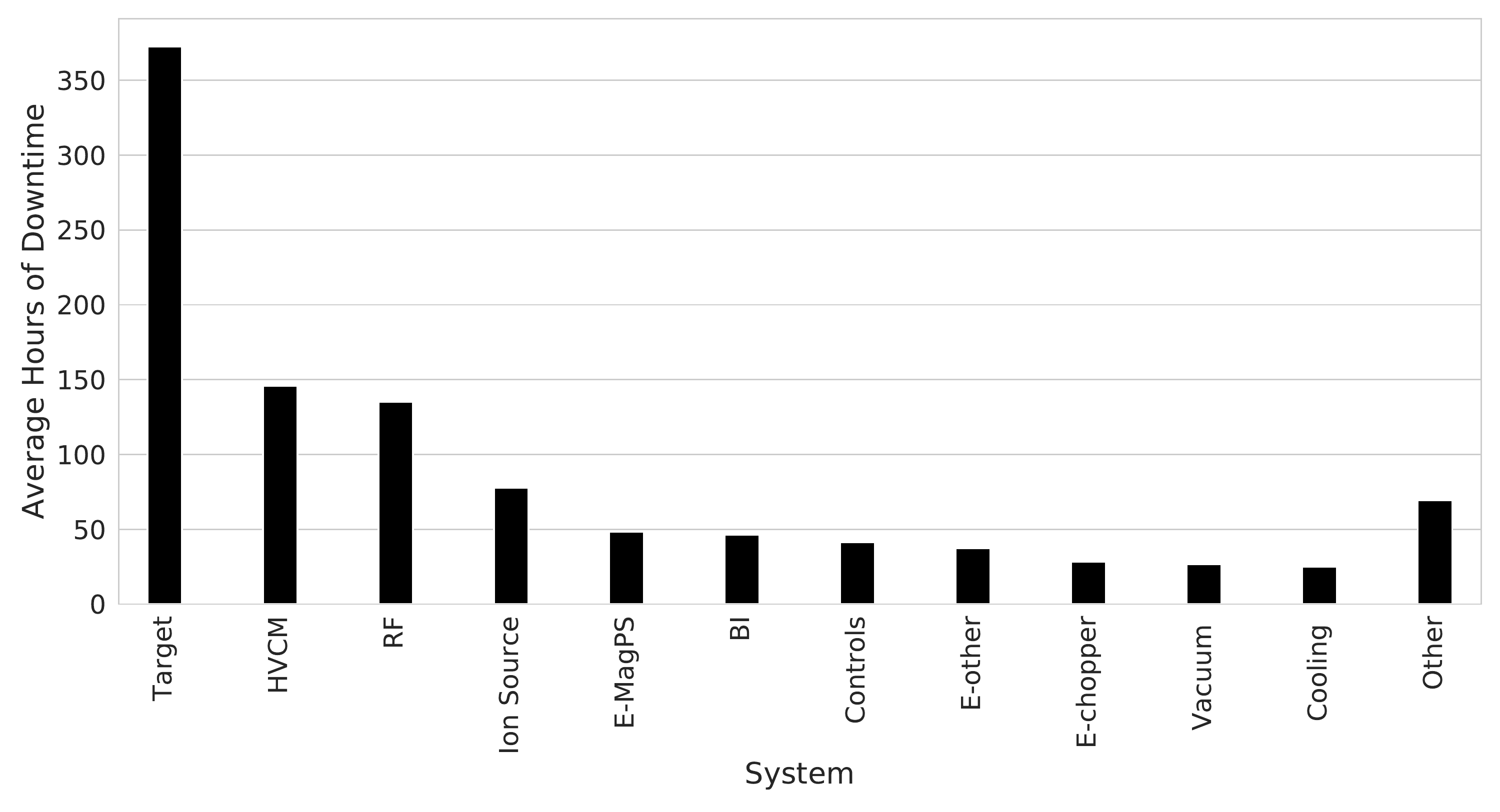}
\caption{SNS Unscheduled Downtime by System. On average, HVCM is the second leading source of downtime after Target. The average is calculated from fiscal year 2007 to 2021.}
\label{fig:History}
\end{figure}

Anomalies or outliers in an engineering system can be caused by multiple reasons including mechanical and human errors. Detecting anomalies and identifying their fault types is critical for the system's normal operation and guide the maintenance to achieve proper functioning of the system. Anomaly detection (AD) has been well studied across a wide spectrum of scientific applications. 
There have been several non-parametric statistical and visualization methods used to detect outliers, including Decision Trees~\citep{John_outliers}, K-Nearest Neighbor~\citep{7251924}, and whisker plot~\citep{Potter2006MethodsFP}. 
Recently, with the advances in Machine Learning (ML), specifically deep learning, much attention has been devoted to detect anomalies using ML techniques. 
ML algorithms have shown significant improvements in detecting outliers for structured and unstructured data.
A comprehensive survey on ML methods to detect anomalies can be found in~\citep{https://doi.org/10.48550/arxiv.1901.03407, 10.1145/1541880.1541882}. 
In a typical AD problem, there is usually a large data set of $normal$ observations, and only a few $abnormal$ observations leading to significant sample bias. Additionally, the limited available $abnormal$ observations might not include all types of potential anomalies. 
As such, a common technique is to develop a ML model that encodes the salient features of the $normal$ data into a reduce representation then decodes this representation back to the original data using an Autoencoder (AE)~\citep{10.5555/2987189.2987190}.
For a well behaved model the decoded data should match the input data and give a small reconstruction error (the difference between the input and decoded data). The assumption for anomaly detection, is that anomalous data will not have the same salient features resulting in a larger reconstructed error, which allows us to set a threshold based on the application requirements to identify $abnormal$ samples. Variational Autoencoder (VAE)~\citep{https://doi.org/10.48550/arxiv.1312.6114} has also been proposed by~\citep{vae_anomaly} to detect anomalies using the reconstruction probability as opposed of using the reconstruction error as in AEs. 
The anomaly score in the proposed approach is calculated in terms of a probability that is extracted by sampling from the mean and variance parameters generated from the probabilistic encoder. Using the reconstruction probability, one can capture not only the difference between the reconstruction and the input data but also the variability of the reconstruction by using the variance parameter of the latent variable. 
AEs and VAEs have been widely used to detect anomalies in multiple scientific applications with different types of data, such as fraud detection~\citep{Adewumi2017ASO}, medical image analysis~\citep{LITJENS201760}, remote sensing~\citep{Ball_2017}, smart manufacturing~\citep{ALFEO2020272}, and Internet of Things (IoT)~\citep{https://doi.org/10.48550/arxiv.1712.04301} where the data are images, and univariate or multivariate time series outlier detection \citep{10.1109/TIP.2017.2713048, ijcai2019-378} where the data are sequences, and video anomaly detection~\citep{https://doi.org/10.48550/arxiv.1510.01553} where the data are sequences of images. Based on the data types, the NNs used in the AE and VAE architectures are different, e.g., Convolutional NNs (CNNs) are typically used for imaging data and Recurrent NNs (RNNs) are usually used for time series data.
\newline
\indent In this paper, we present a study on how to detect anomalies in the High Voltage Converter Modulator (HVCM)~\citep{1288975} at the Spallation Neutron Source (SNS) facility~\citep{SNS_cite} in order to reduce long downtimes.
The HVCMs consist of multiple modules working cooperatively to produce high quality neutron beams at the SNS. Therefore, it is critical to detect faults ahead of time to avoid long downtime. As shown in Figure~\ref{fig:History}, on average the HVCM is the second largest source of downtime from fiscal year 2007 to 2021 at the SNS. While the scope of this paper is limited to the HVCM system, it is worth mentioning that there is concurrent work by other collaborators dedicated to reduce the SNS downtime caused by Target~\citep{radaideh2022model}, which is the leading source of downtime at SNS. Previous studies using ML methods have been used to detect anomalies in the HVCMs, such as,~\citep{pappas2021machine, radaideh2022time, radaideh2022real}. These studies showed promising results for detecting faults for a single module using a single waveform, however, the HVCMs consist of 15 modules with 14 sources of waveforms that were not considered in the previously developed ML model. In this study, we include all waveforms and HVCM module types. We found that different waveforms are sensitive to different fault types, and using a single waveform is only sufficient to detect anomalies coming from the corresponding source of waveform. 
For example, a fault in A-FLUX waveform (magnetic flux in the A-phase of the HVCM), might not be detected in MOD-V waveform (modulator voltage) and vise versa. We also found that it is more efficient to incorporate all modules instead of using a single ML model for each file. 
This is because including all modules enforces the model to learn diverse representation of $normal$ data, and hence, can generalize well to several anomalies.
\newline
\indent In this study, we present our results using a multi-module Conditional Variational Autoencoder (CVAE) model to detect anomalies. 
The model was trained using all 14 source waveforms for each HVCM module and all 15 HVCM modules.
By using a CVAE and all 15 HVCM modules we increase the overall number of samples which can improve the model performance for modules with limited samples, eliminates the need of retraining a model for each module individually, and allow us to develop a well-performed AD model that can generalize well to various anomalies.
\newline
\indent We evaluate our trained models by visualizing their loss landscapes using filter normalization technique~\citep{NEURIPS2018_a41b3bb3} to assist in hyper-parameter optimization and model selection to produce well-performed predictions. We present a side-by-side loss landscape comparisons between the proposed multi-module with a single-module that is trained individually for each HVCM module. The results of this analysis indicate that the multi-module approach can learn from multiple modules and produces a convex-like loss space for all HVCM files, while the single-module model has chaotic loss surface for some HVCMs using a fixed set of parameters and NN architectures
\newline
\indent The rest of this document is organized as follows. 
In Section~\ref{ch:hvcm}, we provide an overview of the HVCM. Section~\ref{ch:previous} reviews the previous work to detect anomalies in the HVCMs at the SNS. 
Section~\ref{ch:background} gives a brief background of AE, VAE, and CVAE. The experimental data is presented in Section~\ref{ch:method}. Section~\ref{ch:res} demonstrates the results and the loss landscape analysis. The conclusion is finally presented in Section~\ref{ch:conclusion}. 
\section{High Voltage Converter Modulator (HVCM)}
\label{ch:hvcm}
\begin{figure}
\centering
\includegraphics[width=\textwidth]{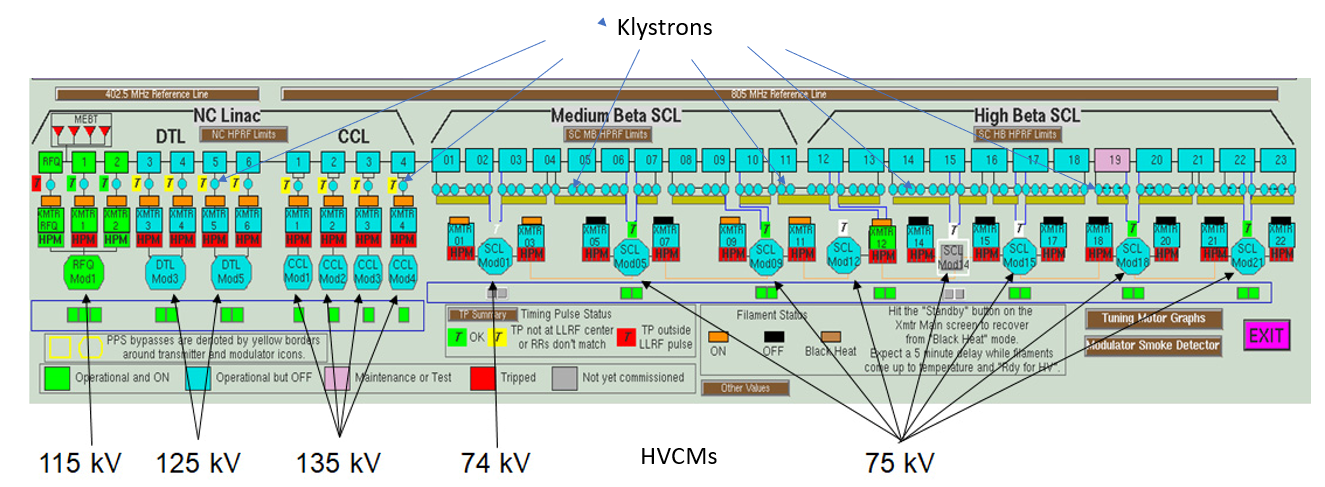}
\caption{Layout of the RF systems at the SNS showing the HVCMs.}
\label{fig:SNSRF}
\end{figure}

The SNS facility, which is the world’s highest power pulsed spallation source, consists of a linear accelerator (LINAC), accumulator ring and transfer line to deliver protons to a target used to produce neutron beams. The SNS LINAC, as shown in Figure~\ref{fig:SNSRF}, is comprised of several different types of radio frequency (RF) structures used to accelerate the beam from low energy to high energy, starting with the normal conducting (NL) cavities and ended with superconducting (SCL) ones. At the lowest energy, a radio frequency quadrupole (RFQ) is used followed by six drift tube LINAC (DTL) cavities. The remainder of the NL Section uses four coupled cavity (CCL) structures for acceleration, while the higher energy Sections use 81 superconducting RF cavities. Each of the cavities is powered by a high power RF amplifier, or klystron~\citep{Caryotakis2004HighPK}. 
Since different types of RF cavities at the SNS operate at different frequencies and require different RF power, they are powered by different types of klystrons which require different cathode voltages and appear as different impedances to the high voltage power supply. SNS uses pulsed klystrons, and the HVCMs are the pulsed power supplies used to drive the klystron’s cathodes. The HVCMs were designed to operate at approximately 1 MW of average power, so each of the HVCM modules powers differing numbers of klystrons in order to use the minimum number of HVCMs. This results in slightly different designs of the HVCMs in order to accommodate the different voltages and load impedances for the klystrons. The types of HVCMs are those for the RFQ and first two DTL Sections, the other DTL klystrons, the CCL Section and the SCL Section.
\section{Previous Work} 
\label{ch:previous}
Although the anomaly detection efforts for the HVCM are limited, there are a few recent studies demonstrated the promise of using ML for anomaly detection in the HVCM systems. 
\citep{pappas2021machine} conducted anomaly detection on one of the SNS HVCMs - SCL09 (super-conducting LINAC) - using discrete cosine transform. Given the work was published in the early phase of the project, very limited data was available, as the authors have used 83 waveform samples (all are magnetic flux in the B-phase). 
Nevertheless, the authors were able to detect 11 of the 16 fault events. \citep{radaideh2022time} extended the effort by applying deep learning RNNs on the RFQ module (radio-frequency quadrupole), the module that had a significant number of failures in the SNS. 
The authors developed long short-term memory (LSTM), gated recurrent unit (GRU), and convolutional LSTM (ConvLSTM) autoencoders, while using the C-FLUX waveform for anomaly detection. 
The authors demonstrated promising results by detecting 39 out of 50 fault events with a false positive rate of about 10\%. 
It is worth highlighting that both of these efforts~\citep{pappas2021machine, radaideh2022time} highlighted a single HVCM module and a single waveform, whereas this study extends the approach to multiple waveforms (multivariate) and multiple HVCM modules (multi-system). 
Implementing this multi-system AD study is possible thanks to a larger dataset collected over two years, which includes many fault events in all 15 HVCM modules of the SNS \citep{radaideh2022real}.
Anomaly detection applications in particle accelerators for systems other than the HVCM have also been demonstrated. AEs are based on vanilla feed-forward NNs were developed for AD of magnet faults in the Advanced Photon Source storage ring at Argonne National Laboratory~\citep{edelen2021anomaly}. 
While the developed models demonstrated a good ability to learn the binary output of whether a fault would occur or not, the authors concluded that their models were unsuccessful in accurately predicting the timing of the fault~\citep{edelen2021anomaly}.  
Unsupervised ML techniques with feature extraction support were applied to exploit the data from a RF tuning system in the ALPI accelerator at Legnaro National Laboratories in Italy \citep{marcato2021machine}. 
The authors of \citep{rescic2020predicting} employed different ML binary classifiers (e.g. logistic regression, gradient boosting, random forests~\citep{breiman2001random}) to predict machine failures via beam current measurements before they actually occur. 
The models achieved failure prediction accuracy up to 92\%. The binary classifiers were then improved in a subsequent study by the team \citep{revsvcivc2022improvements} for preemptive detection of machine trips in the SNS using differential beam current monitor (DCM); achieving a precision of 96\% with 58\% true positive and 0\% false positive rates \citep{revsvcivc2022improvements}. The work by \citep{edelen2018opportunities} highlighted opportunities of ML for various applications in particle accelerators including but not limited to anomaly detection, machine protection, system modeling, diagnostics, tuning, and system control. 
\citep{PhysRevAccelBeams.25.122802} has proposed an uncertainty aware ML method using Siamese model~\citep{Koch2015SiameseNN} to predict upcoming errant beam pulses from a single monitoring device at the SNS.  
These results showed an approximately 2x improvement in identifying anomalies over the previous published results in the region of interest.
\section{Background}
\label{ch:background}
\subsection{Autoencoders}
An autoencoder (AE) is a type of NN for unsupervised learning that has two components: an $encoder$ and a $decoder$. 
As shown in Figure \ref{fig:AE_VAE} (a), the $encoder$ denoted by $\Psi$ learns to encode original input data $x$ typically into a lower dimensional latent space (bottleneck) using a NN. 
The encoded representation is then passed to the $decoder$ denoted by $\Phi$ that learns to reconstruct the original input data using a NN.
The goal of training an AE is to select $\Psi$ and $\Phi$ functions that have the minimal error to reconstruct the input data. 
The loss function used to train an AE is called $reconstruction~loss$, that is a comparison of how well the output has been reconstructed from the original input. 
The $reconstruction~loss$ of a typical AE can be defined as,
\begin{equation}
    \mathcal{L}_{\rm AE}=\|x-\Phi(\Psi(x))\|^2
\end{equation}
that takes the difference between the original input data and the reconstructed one, where $x$ is the input data. 
There are different metrics proposed in the literature to be used as a $reconstruction~loss$; one such example is Mean-Squared-Error as shown in Equation~\ref{eq:1}, where $x$ is the input data and $\hat{x}$ is the reconstructed data from the model.
\subsection{Variational Autoencoders}
\begin{figure}[!tbp]
  \centering
  \subfloat[AE Architecture]{\includegraphics[width=0.45\textwidth]{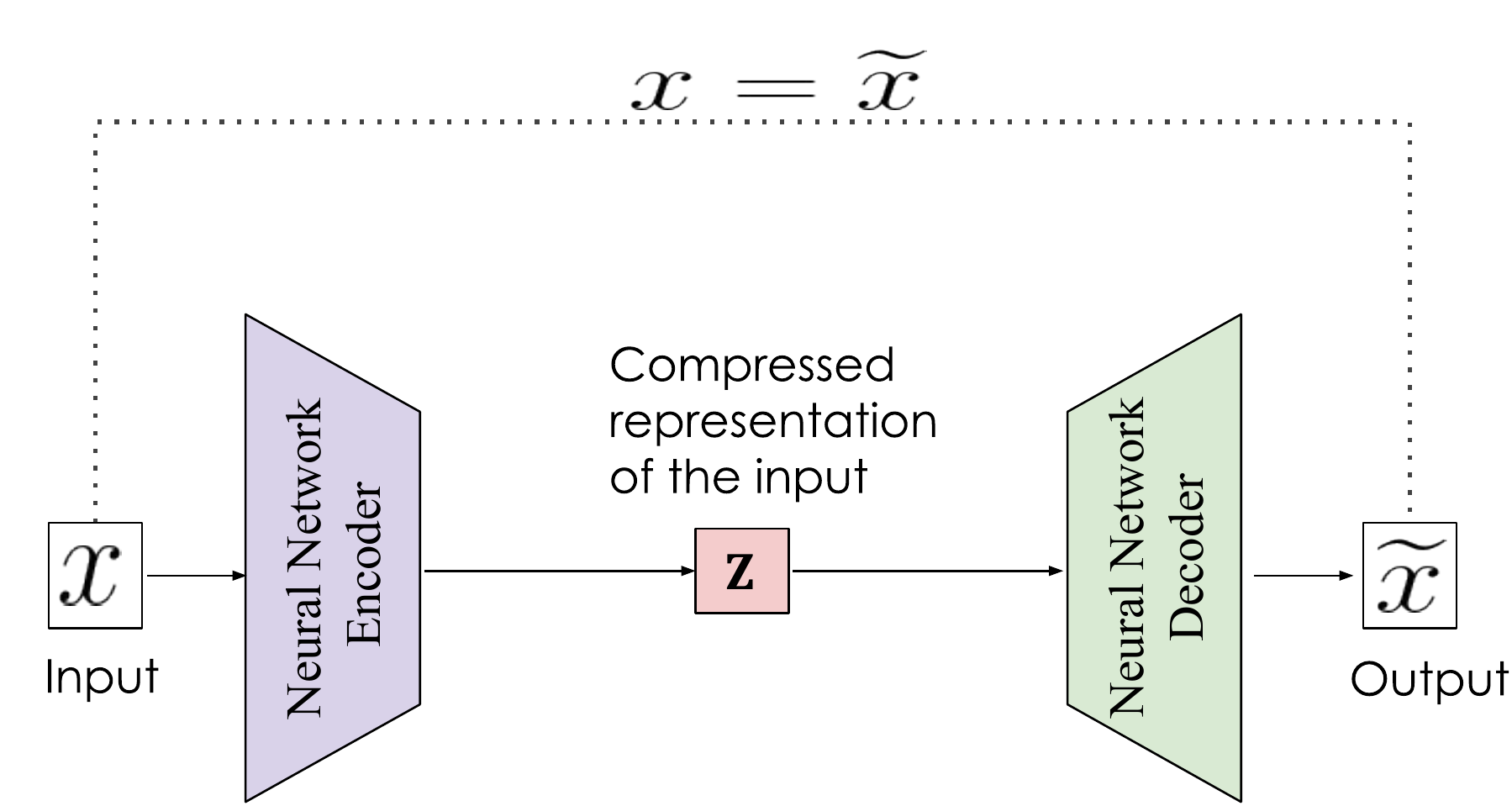}\label{fig:f1}}
  \hfill
  \subfloat[VAE Architecture]{\includegraphics[width=0.45\textwidth]{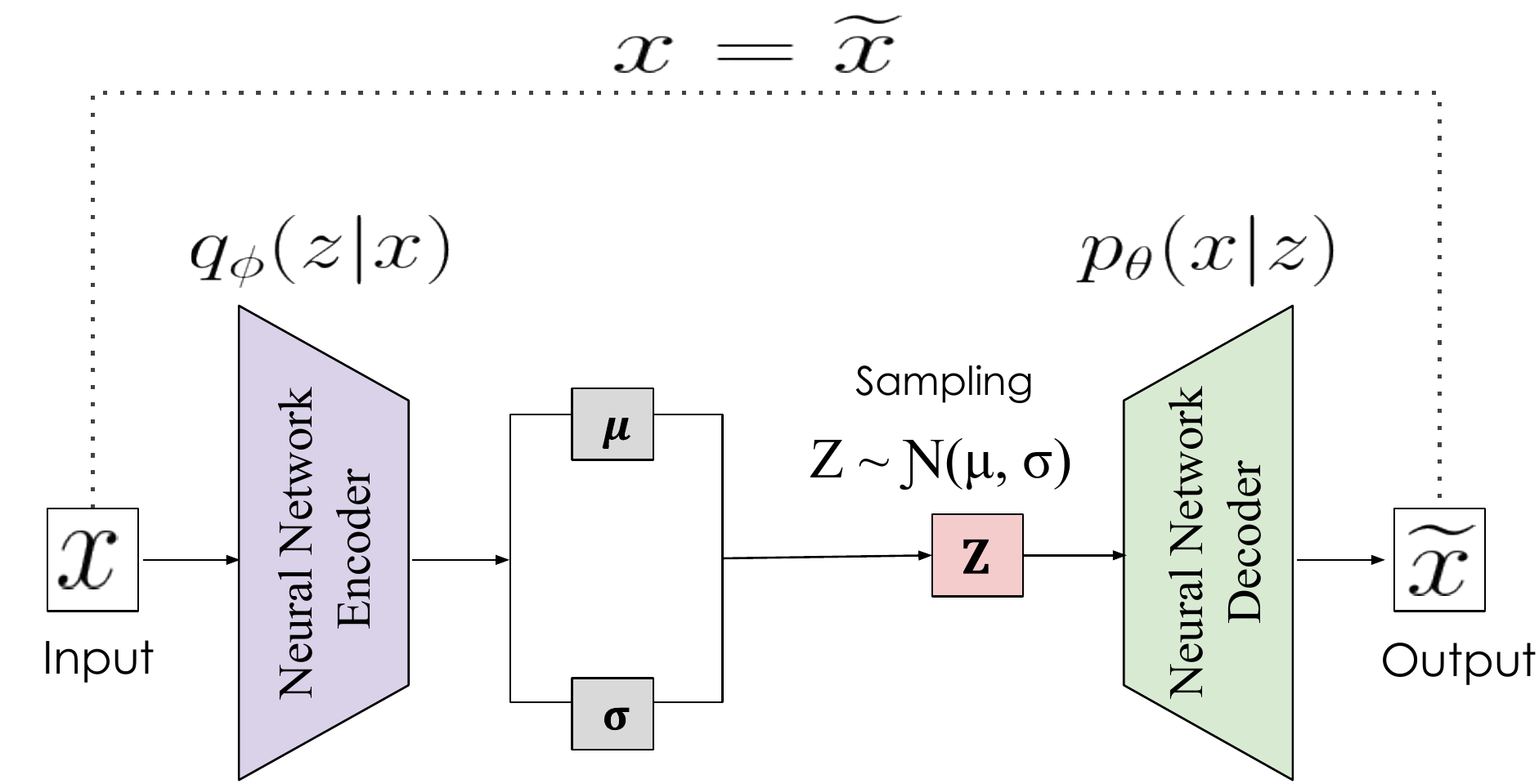}\label{fig:f2}}
  \caption{Figure (a) shows a typical Autoencoder (AE) model that consists of an $encoder$ that projects input data into smaller representation $z$, and a $decoder$ that reconstructs input data given latent $z$. Figure (b) shows a typical Variational (VAE) model consists of an $encoder$ that projects the input data into a probability distribution and estimates the $\mu$ and $\sigma$ parameters of that distribution. The sampling layer $z$ takes the estimated parameters to sample a distribution that needs to be as close as possible to the pre-defined distribution. The $decoder$ takes the sampling layer $z$ as inputs to reconstruct the input data and generate new samples.}
  \label{fig:AE_VAE}
\end{figure}

While an AE learns a function to map each input to a fixed size single point reduced representation in the latent space, a Variational Autoencoder (VAE) replaces the latent space with probability distribution of the input data by replacing the (bottleneck)
with two separate vectors $\mu$ and $\sigma$, representing the mean and the standard deviation of the distribution. As shown in Figure \ref{fig:AE_VAE} (b), the $encoder$ $\Psi$ projects high-dimensional input data $x$ into a lower latent variable $z$ that is forced to follow a certain well-known distribution, such as a Gaussian. The latent variable $z$ in VAE is not the output of the $\Psi$ as in AE, instead, the $\Psi$ estimates $\mu$ and $\sigma$ parameters for each latent variable. Then, the latent $z$ is sampled from the estimated parameters which is fed to the $\Phi$ to reconstructs the input data. A new parameter $\epsilon$ is also introduced to allow us to reparametrize the sampling layer $z$ and allow the model to backpropagate the entire network. 
The latent $z$ is now defined as, $z= \mu +\sigma\odot\epsilon$, where $\epsilon\sim\mathcal{N}(0,1)$. A VAE generalizes the idea of an AE by not only learning the embeddings, but also being able to generate new data by sampling from the estimated latent distribution parameters $\mu$ and $\sigma$. The loss function for a VAE is motivated by variational inference~\citep{variationalinference} via minimizing the Kullback–Leibler divergence (KLD)~\citep{kullback1951information} between the posterior $p(z|x)$ and the encoded prior distribution $q(z) = \mathcal{N}(0,1)$:
\begin{equation} \label{eq:2}
    \mathcal{L}_{\rm VAE}= \|x-\Phi(\Psi(x))\|^2 +\eta~{KLD}~(q(z) ~||~ p(z|x))
\end{equation}
where the first term is the reconstruction error, the second term computes KLD, and $\eta$ is the harmonic parameter to balance the two.

\subsection{Conditional Variational Autoencoders}
For structured output predictions, Conditional VAEs (CVAE) were proposed by~\citep{NIPS2015_8d55a249} to make diverse predictions for different input samples. The objective function of the VAE can be modified by adding the variable $c$,
\begin{equation} \label{eq:3}
    \mathcal{L}_{\rm CVAE}=\|x-\Phi(\Psi(x))\|^2+\eta~{KLD}~(q(z,c)~ ||~ p(z|x,c))
\end{equation}
where we condition all of the distributions with 
$c$. CVAEs are an extension of VAEs by adding the conditional part in the $encoder$ and $decoder$ to associate the input samples with labels.
Therefore at inference time, we have more control to generate samples that belong to specific labels in contrast to a VAE that does not have control over the generated samples. 
\section{Methods}
\label{ch:method}
In this section, we will first describe the experimental data used in this paper, then we will describe the architecture of the single-module based VAE and the multi-module based CVAE.
\subsection{Data Description}
\label{DATA:lab}
\begin{figure}
\includegraphics[width=\textwidth]{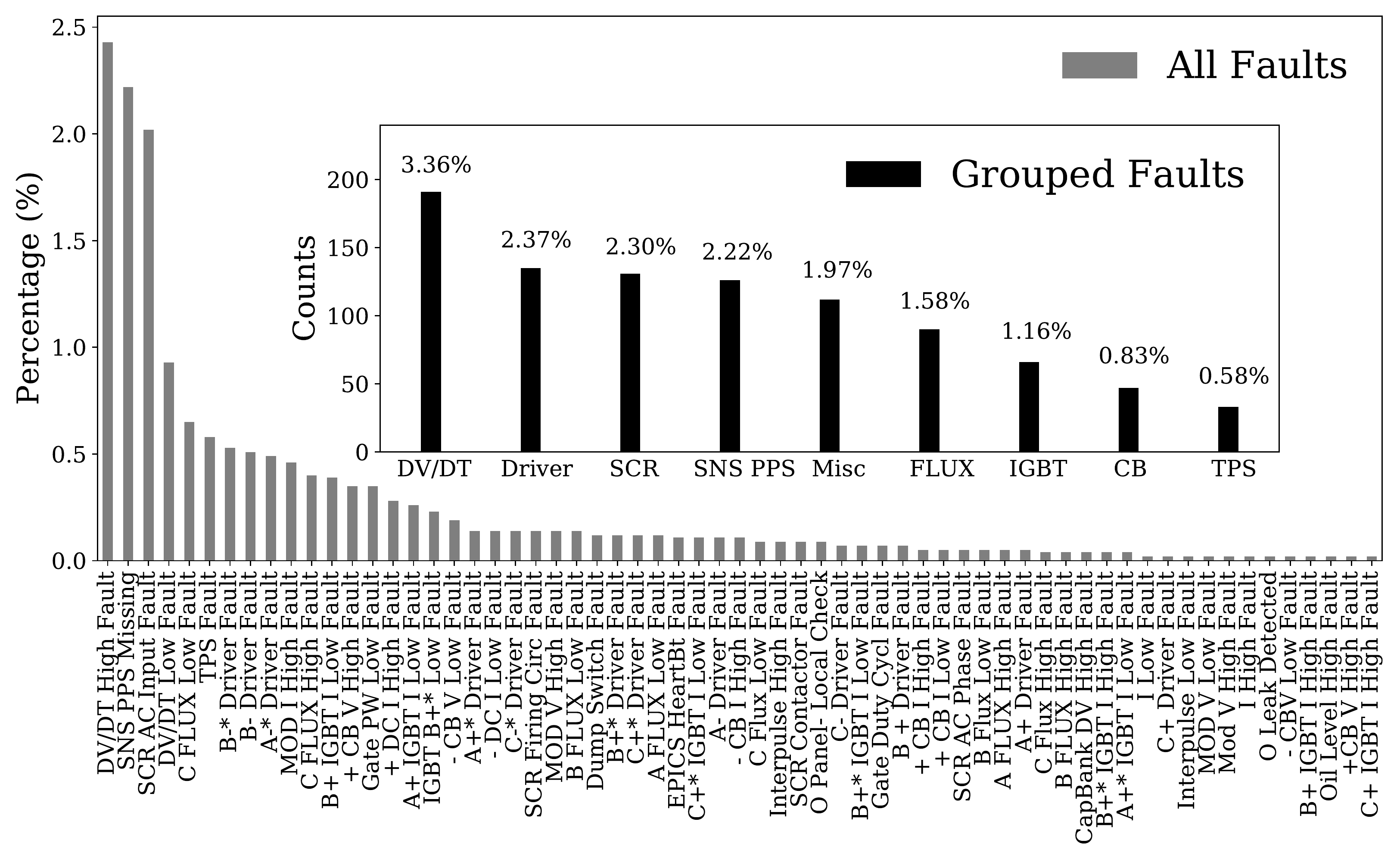}
\caption{The outer figure (grey bars) shows percentage of $abnormal$ waveform types with respect to all data including $normal$. The inner figure (black bars) shows the counts of $abnormal$ data after regrouping.}
\label{fig:fault}
\end{figure}

We train and test our developed models on experimental data extracted from the HVCM controller.
Historically, the IGBT switches (Insulated-gate bipolar transistor) have been a significant source of unplanned downtime, and still occasionally have catastrophic failures which could result in several days of lost neutron production. 
Reliability has significantly improved over the years with upgrades to the HVCM. 
The upgrades provided a rich source of digitized waveforms from the HVCMs which will be used for ML. 
The control of the HVCMs is done with a PXI-based controller running LabView. 
Along with providing communications and control to the HVCMs, the controller digitizes up to 32 waveforms at a sample period of 20 ns. 
The waveforms are saved to an on-board computer with a record length of 3 ms. 
The controller also saves another file which is decimated to 2.5 MS/sec but with a record length of 35.5 ms to capture ``three'' macro-pulses. 
These files are overwritten every macro-pulse with the exception of when the HVCM faults, and after maintenance has been performed or the system re-tuned in order to record $normal$ waveforms. 
The controller saves a file containing all of the HVCM settings including configuration parameters.
From the 32 waveforms saved by the controller, we only used 14 waveforms in this work based on expert opinion on their importance to HVCM reliability. 
From the full record length of 35.5 ms, we extracted 1.8 ms macro-pulses according to the following strategy:
\begin{itemize}
    \item For $normal$ waveforms, all three macro-pulses are extracted since they are identical and can increase the number of data samples. Each 1.8 ms $normal$ macro-pulse has 4500 time steps (i.e. sampling rate is 400 ns)
    \item For faulty waveforms, the pre-fault pulse  was used to predict the anomalies (i.e. detecting the fault ahead of time to allow system trip). Typically in faulty waveform files, the first macro-pulse comes before the fault event (pre-fault), the second macro-pulse has the fault, while the third macro-pulse is usually not saved as the HVCM is down. The faulty 1.8 ms macro-pulse also has 4500 time steps (i.e. sampling rate is 400 ns).
\end{itemize}
These data pre-processing strategies were done by the team before in their previous work \citep{radaideh2022time}, primarily to reduce data size and cut the irrelevant time steps when the HVCM is idle. 
See Figure 5, Section 3.2, and Section 4.1 of \citep{radaideh2022time} for more information about data processing. The processed data used in this work is shared with the public \citep{radaideh2022real}.  
The 14 waveforms (features) used in this work are: 
\begin{itemize}
    \item Six IGBT current waveforms, which express the current passing in the phases A+, A+*, B+, B+*, C+, and C+*.
    \item Three magnetic flux density in the phases A, B, and C of the resonant circuit. 
    \item Two waveforms that represent the cap bank voltage and the cap bank current. 
    \item Two waveforms that represent the modulator output voltage and the modulator current. 
    \item One waveform that represents the time derivative change of the modulator output voltage (i.e. dV/dt).
\end{itemize}
After data processing, the number of $normal$ and $abnormal$ samples are saved in a 3D tensor of shape:
(5240, 4500, 14) and (1270, 4500, 14) respectively, where the time steps is 4500 and the waveforms features are 14 as described above. 
The $abnormal$ waveform types before and after grouping are shown in Figure~\ref{fig:fault}.
\subsection{Anomaly Detection Approach}
\begin{figure}[h]
\includegraphics[width=\textwidth]{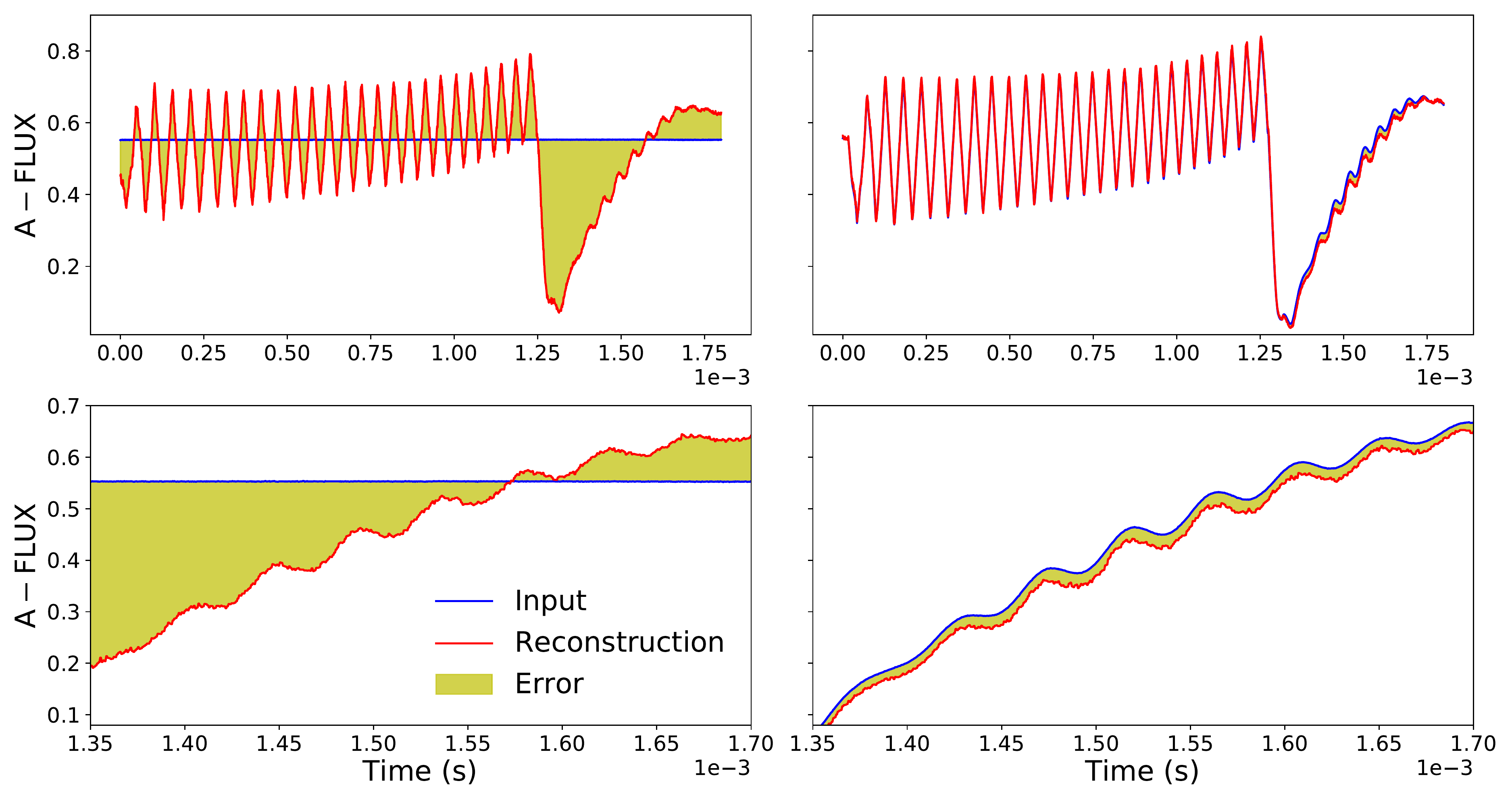}
\caption{Top row shows two different observed faulty waveforms (blue) and their corresponding reconstructed samples from multi-module CVAE (red); the difference between the observed and reconstructed waveform is highlighted in yellow. Bottom row shows a subset of the waveforms presented on the top row to zoom in a specific region.}
\label{fig:wave}
\end{figure}

To detect anomalies using VAEs, we train the model using $normal$ waveforms to capture the normal behavior. Afterwards we use the trained model to detect anomalies by calculating the difference between the input waveform with the predicted waveform using Equation~\ref{eq:1}, commonly known as the $reconstruction~error$. 
The $normal$ waveforms are expected to have smaller $reconstruction~error$ than $abnormal$ waveforms, therefore, we can define a threshold to achieve the desired requirements.
For this paper, we require less than 10\% false positives in order to compare to previously published results. For example, in Figure~\ref{fig:wave} we have two different $abnormal$ waveforms (blue color) and the reconstructed waveforms from our model (red color), with the $reconstruction~error$ (yellow band) measuring the difference between the input and reconstructed waveforms. 
We can see the input $abnormal$ example on the left is flat and can be easily identified as faulty even without the need of any ML techniques; however, these easily identified faulty waveforms represent a small fraction of the $abnormal$ data, whereas most of the other faulty examples are very similar to $normal$ data, and cannot be identified easily, similar to the right plot of Figure~\ref{fig:wave}. 
The bottom row of the Figure~\ref{fig:wave} shows a subset from the top waveforms to zoom in a specific region of the waveforms for clearer visualizations. For faulty waveforms that are difficult to be distinguished from normal ones, ML techniques are needed for detection. The advantage of using VAEs over AEs in this application is that a VAE has a probabilistic model that provides a probability measure instead of a reconstruction error. By using the probabilistic encoder, we can sample from the mean and variance parameters to capture the variability of the reconstruction instead of having a a single reconstruction error as an anomaly score. In the following, we use two ML methods for anomaly detection of HVCM data from SNS. One is a VAE for a single HVCM module, and the other is a CVAE for multiple HVCM modules. 
\subsection{Single-module based Variational Autoencoder}
\label{SINGLE_VAE}
A standard approach when using VAEs on data originating from multiple modules is to train an individual model for each module. 
This allows us to focus on each subsystem independently and detect anomalies specific to the module. 
Initially, we trained separate VAE for each subsystem.
We designed a VAE based on one dimensional convolution layers for both the $encoder$ and a $decoder$. 
The $encoder$ takes waveforms $x$ of shape $(N_{samples}~\rm{x}~N_{time{-}steps}~\rm{x}~N_{features})$ as input that are processed through multiple 1 dimensional convolutional neural network (CNN) layers. 
The output of the last CNN block is flattened and fed to fully connected (dense) layers.
The last dense layer in the encoder produces estimated parameters $\mu$ (mean) and $\sigma$ (standard deviation) representing our prior distribution which is constrained to follow a Gaussian distribution. 
Using these two outputs, $\mu$ and $\sigma$, a customized layer (Lambda) is used to randomly sample a Gaussian distribution that is as close as possible to the prior $q(z)$ to generate the output of the encoder, latent $z$. 
The $decoder$ receives latent $z$ as an input which is then fed to two dense layers in a sequential manner. The resulting output of the Dense layer is reshaped and processed through three 1 dimensional CNN blocks that reconstructs the waveform $x$. 
Similar to a typical CVAE, we use Mean Squared Error (MSE) for the $reconstruction~loss$ and KLD between the posterior $p(z|x,c)$ and the encoded prior distribution $q(z,c) = \mathcal{N}(0,1)$ so the loss function will be identical to Equation \ref{eq:2}. 
The architecture of this model is identical to Figure~\ref{fig:VAE} excluding the conditional part shown in the dashed box. Our multi-module model architecture is inspired by CNNs~\citep{10.5555/303568.303704} and fully CNNs~\citep{NIPS2012_c399862d}. 
To train this model, we use Keras~\citep{chollet2015keras} and TensorFlow~\citep{tensorflow2015-whitepaper} backend with the parameters setup shown in Table~\ref{tab:hyperparam}. 
We create 15 different models, each model simulating one HVCM module.  
We train each model using all the available features discussed in Section~\ref{DATA:lab}. 
After leaving test-out samples, the training data shape becomes $(N~\rm{x}~4500~\rm{x}~14)$, where $N \approx 450$ samples for each subsystem. The results and evaluations of this methodology will be discussed in Section~\ref{ch:res}.
\subsection{Multi-module based Conditional Variational Autoencoder}
\begin{figure}[h]
\includegraphics[width=\textwidth]{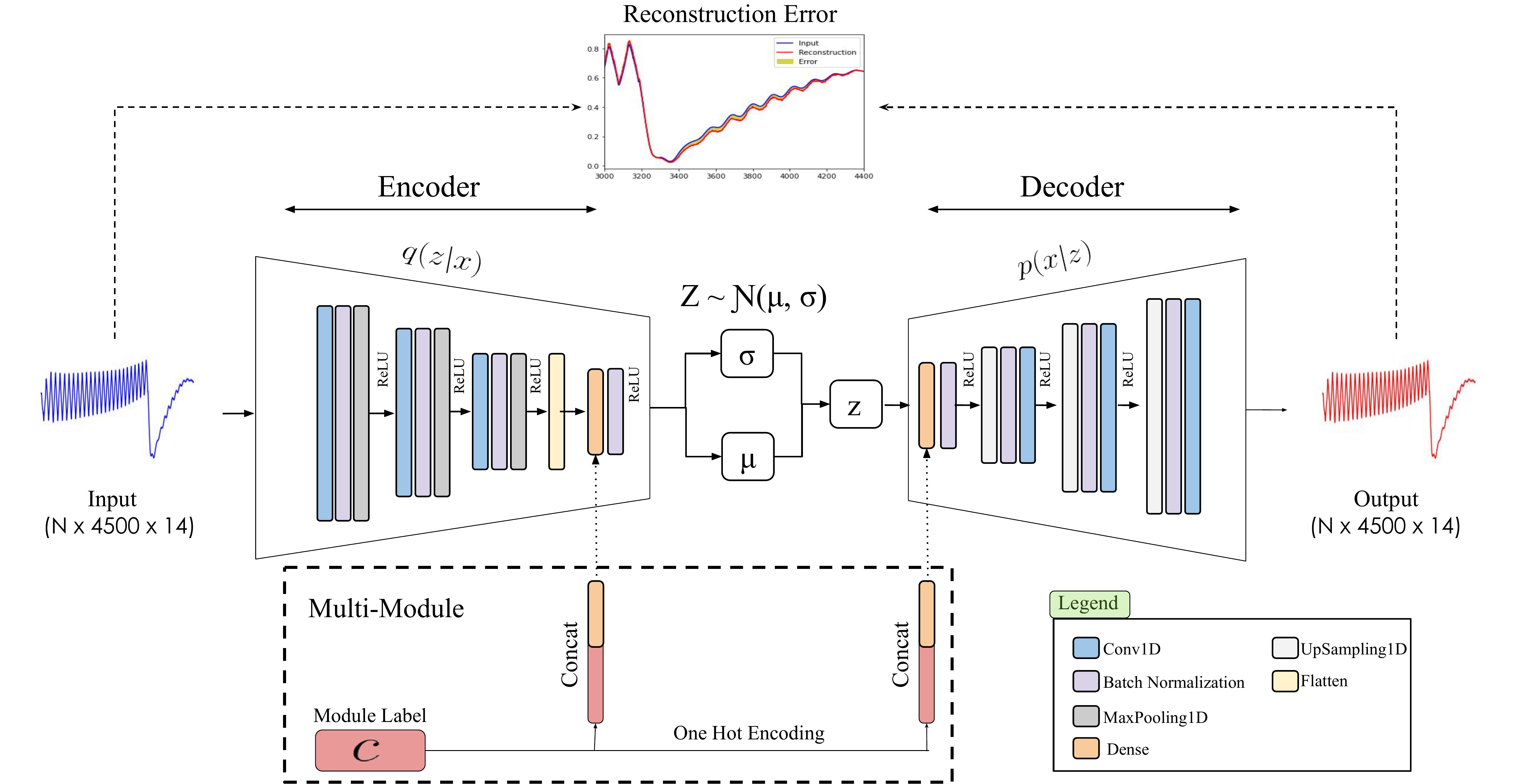}
\caption{Anomaly Detection Method Overview. single-module and multi-module models share the same architecture, but the later has the conditional part shown in the dashed box on the bottom of this diagram.  }
\label{fig:VAE}
\end{figure}

In addition to training an individual model for each HVCM module, we also design a multi-module model that trains all the 15 HVCM systems (discussed in Section~\ref{DATA:lab}) together. 
Motivated by the architecture of CVAE, we extend the single-module based VAE to include a conditional component $c$, which is the unique IDs for different modules.
In Figure~\ref{fig:VAE}, we show the architecture of our model where we have an $encoder$ that takes $normal$ waveforms $x$ of shape $(N_{samples}~\rm{x}~N_{time{-}steps}~\rm{x}~N_{features})$ as inputs, that are passed through three 1 dimensional CNN blocks. 
The output of the last block is flattened and fed to a fully connected layer concatenated with One-Hot-Encoding of the module unique IDs, which then is passed to another fully connected layer that will generate the latent $z$ which is constrained to follow a Gaussian distribution, providing an estimated parameters $\mu$ and $\sigma$. 
The $decoder$ receives latent $z$ and the same One-Hot-Encoding of the module unique IDs as inputs which are concatenated and fed to two dense layers in sequential order. 
The output of the fully connected layers is reshaped and fed to three 1 dimensional convolutional blocks that will reconstruct the waveforms $x$. 
Similar to the previous model, we use MSE for the $reconstruction~loss$ and KLD between the posterior $p(z|x,c)$ and the encoded prior distribution $q(z,c) = \mathcal{N}(0,1)$ so the loss function will be identical to Equation~\ref{eq:3} after adding the conditional component. 
The idea of combining all 15 HVCM modules together is to allow the model to learn different representations of $normal$ waveforms, improve the model performance
for modules with limited samples, eliminates the need for individually retraining each module model and to provide a more generic solution than using a single-module based model. 
By adding the conditional component $c$, the model can learn the association between the waveforms and their modules.
To train this model, we use all available 15 modules and features discussed in Section~\ref{DATA:lab} that have 5240 waveforms for each feature. 
After leaving test-out samples, the training data shape becomes $(524~\rm{x}~4500~\rm{x}~14)$. 
We use the same parameters configuration as in the previous model which is shown in Table \ref{tab:hyperparam}
\begin{table}
\centering
\begin{tabular}{p{5cm} | p{4cm}}
\hline
\hline
Parameter & Value\\ [0.2ex] 
\hline
Input/output dimensions & (N,4500, 14)\\
$encoder$ Conv1D blocks & 3\\
$decoder$ Conv1D blocks & 3\\
Number of kernels & 128\\
kernel size & 3\\
Activation function & $ReLU$\\
Unites per Dense layer & 512\\
Latent $z$ & 512\\
Optimizer & Adam\\
Batch Size & 16\\
Loss & $MSE+KLD$\\
Learning rate & $10^-5$\\ 
\hline
\end{tabular}
\caption{\label{tab:hyperparam}Model Setup.}
\end{table}
\section{Results}
\label{ch:res}
\begin{figure}
  \centering
  {\includegraphics[width=0.49\textwidth]{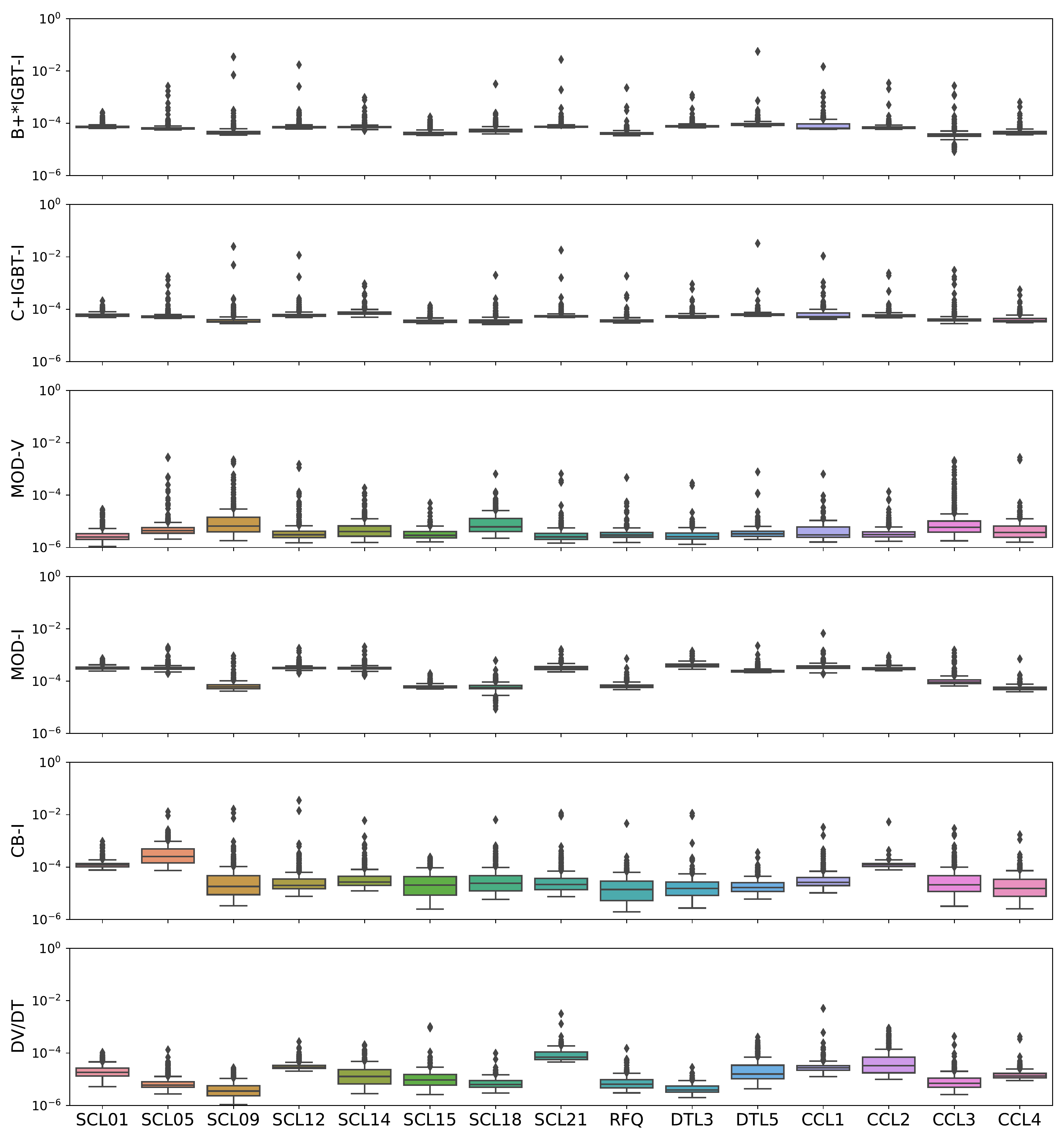}\label{fig:box1}}
  \hfill
  {\includegraphics[width=0.49\textwidth]{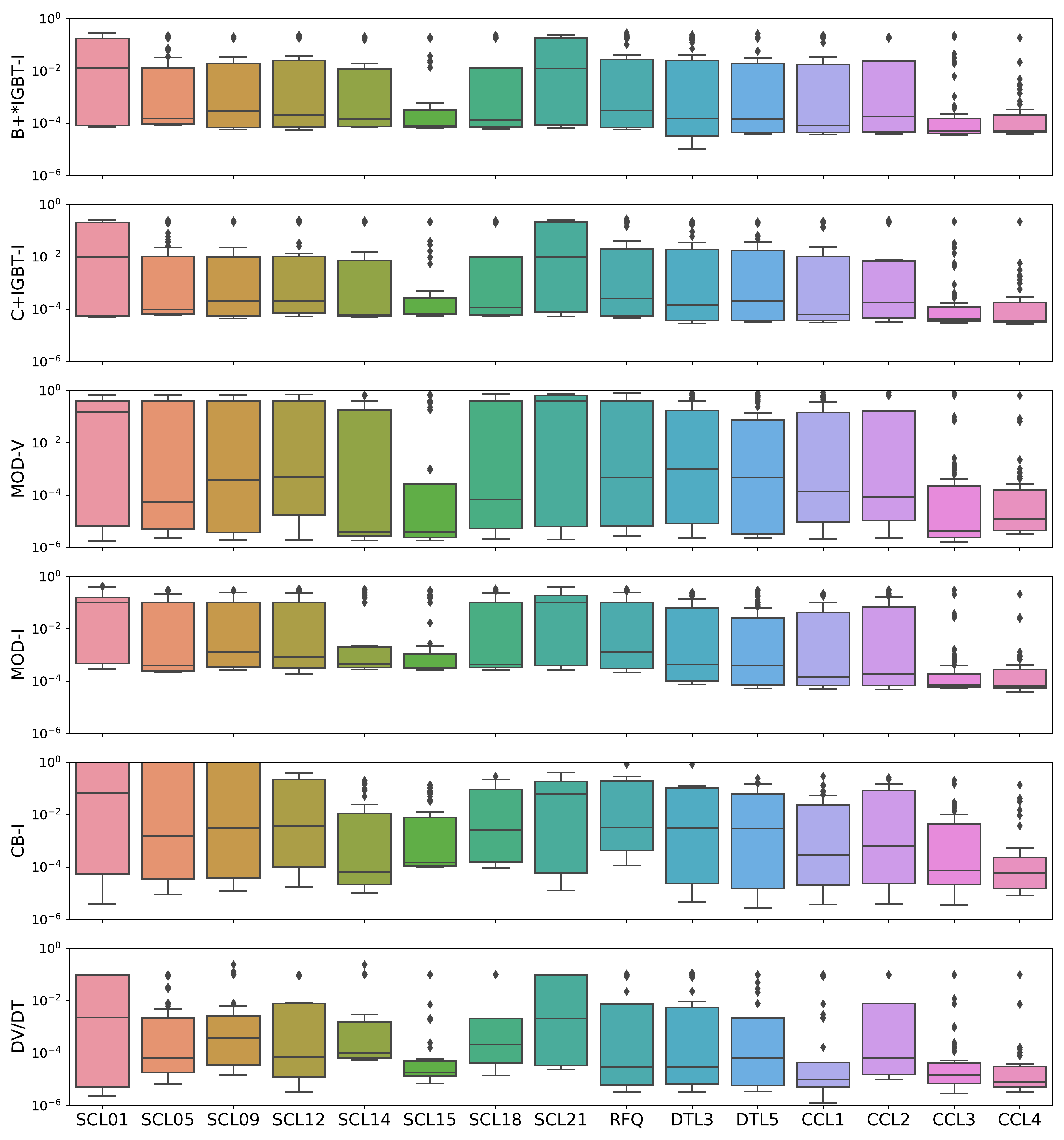}\label{fig:box2}}
  \caption{Box plot shows the MSE $reconstruction~error$ distributions of $normal$ (left plot) and $abnormal$ (right plot) using multi-module. The x-axis shows all trained modules and the y-axis shows six source of waveforms.}
  \label{fig:boxplot}
\end{figure}

For this work, we use several evaluation techniques to quantify the performance of two methodologies. 
The first is the multi-module that combines all systems together, while the second is the single-module that is trained individually for multiple systems. 
We first show the results for the multi-module model and discuss the accuracy of the detecting anomalies, then we compare the two models and have side-by-side plots. 
In Figure~\ref{fig:boxplot}, we show the MSE $reconstruction~error$ distributions between $normal$ and $abnormal$ data of multi-module model using box plots. 
Figure~\ref{fig:boxplot} shows the ability of the model to produce very small error when reconstructing $normal$ waveforms, while the model tends to generate larger errors for $abnormal$ waveforms. 
We can see in almost all waveforms sources, such as MOD-V and MOD-I, only using the mean of the box plot we can have a threshold to separate $normal$ from $abnormal$ waveforms.

\begin{figure}[h]
\includegraphics[width=\textwidth]{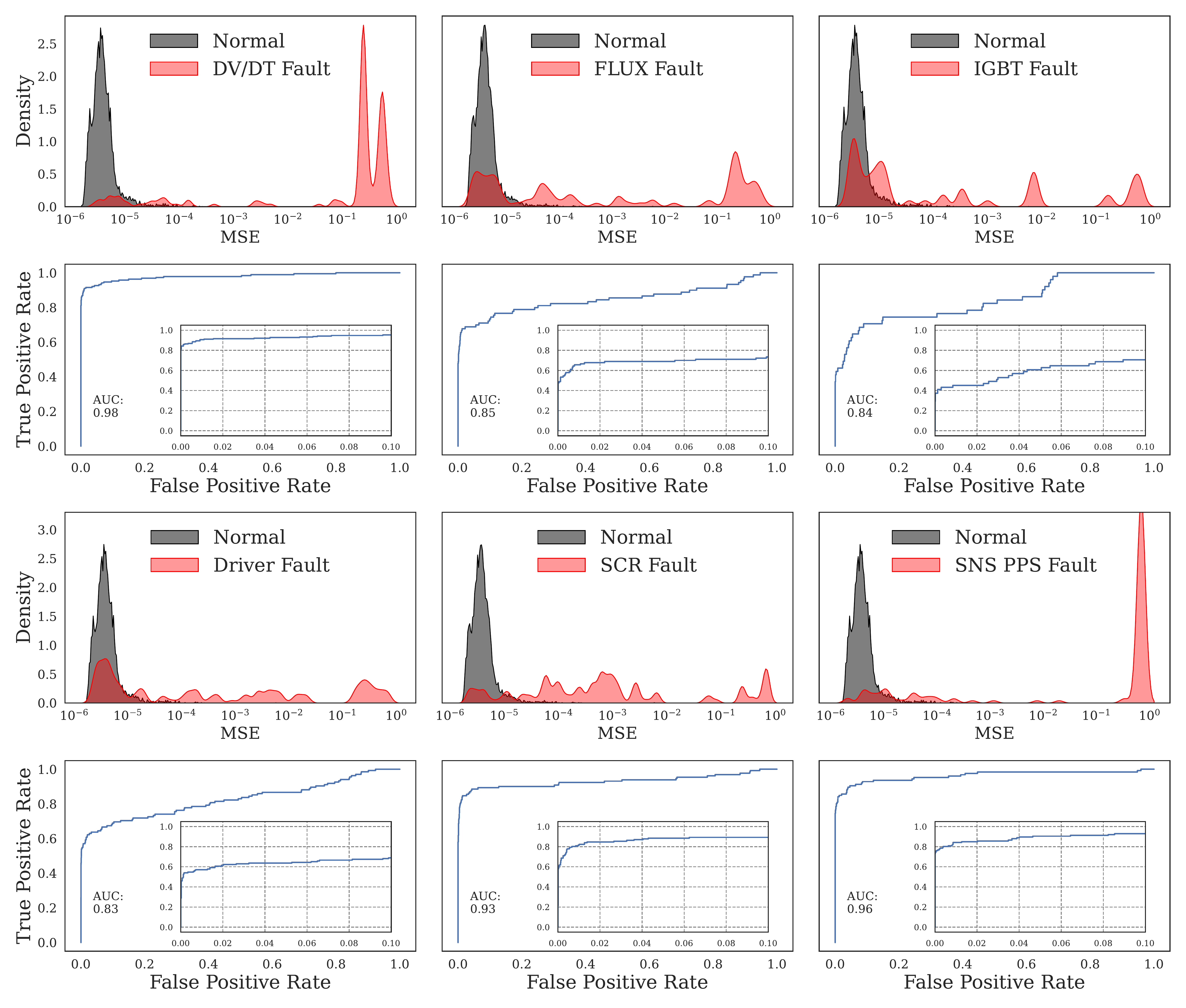}
\caption{KDE distributions of the MSE from reconstructing normal (grey color) and faulty waveforms (red color) for six fault types, with the corresponding ROC curve for each fault.}
\label{fig:multi}
\end{figure}

\begin{figure}
\includegraphics[width=\textwidth]{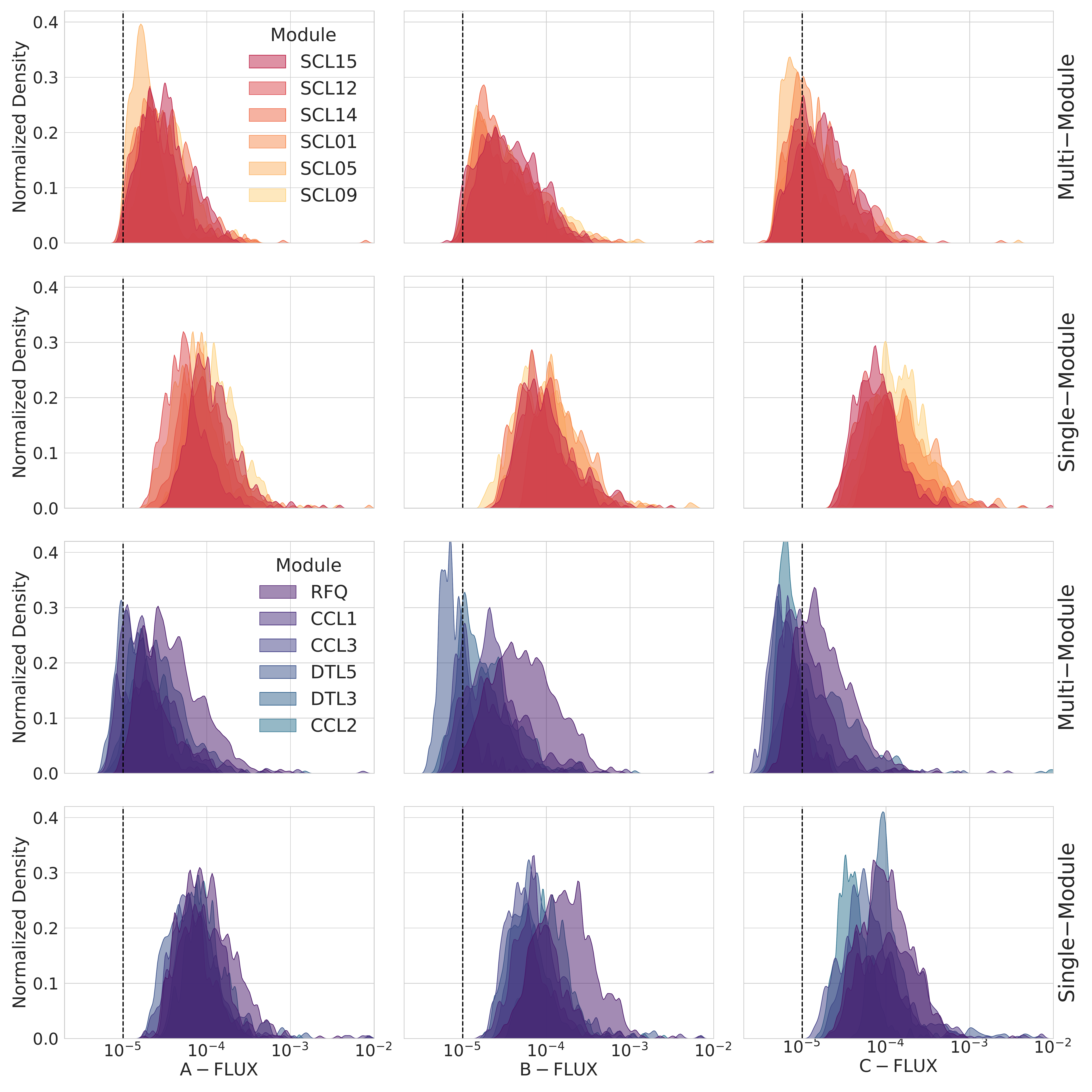}
\caption{Normalized density estimation plot shows the reconstruction error (MSE) of normal waveforms using multi-module and single-module. The multi-module model shows smaller reconstruction error, where the distributions of most of the individual systems are more shifted to the dashed black line at $MSE=10^{-5}$.}
\label{fig:normal_mse}
\end{figure}

\begin{figure}
\includegraphics[width=\textwidth]{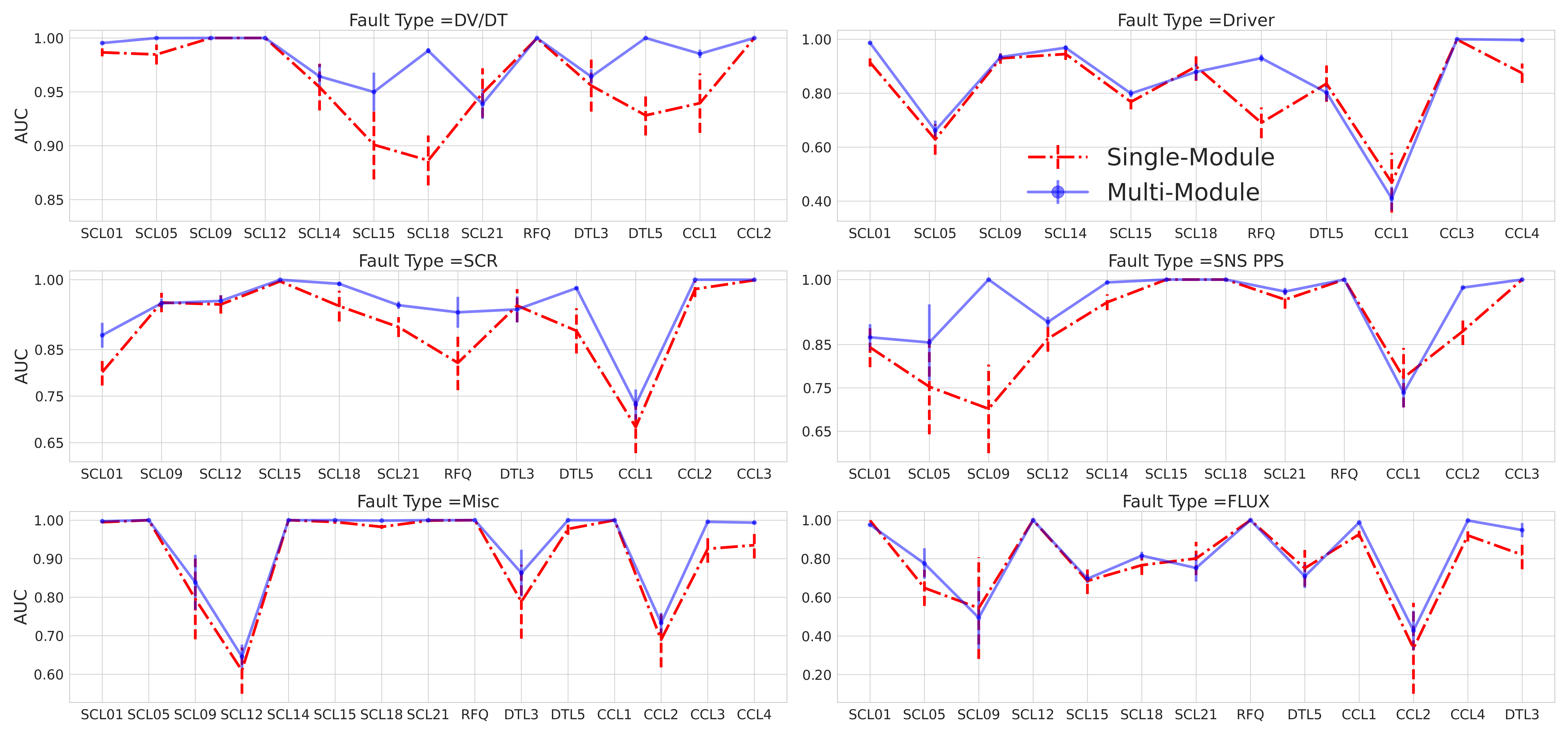}
\caption{Compare the AUC values between single-module and multi-module using six types of faults across several modules. The error bar is $\pm$ 1 Standard Deviation (SD) error generated by sampling the latent $Z$ of each Method.}
\label{fig:auc}
\end{figure}

Next, in Figure \ref{fig:multi}, we show a kernel density estimate (KDE) plot, and the corresponding Receiver Operating Characteristic (ROC) curve with the Area Under The Curve (AUC) values using the $reconstruction~error$ from the multi-module model. 
For this, we use six faults: (DV/DT, FLUX, IGBT, Driver, SCR, and SNS PPS faults) and use the $reconstruction~error$ from MOD-V waveforms that produced the highest AUC values. 
We can see that for certain fault types, such as DV/DT and SNS PPS, the model is capable of separating $normal$ waveforms from $abnormal$ waveforms, where the density estimation for $normal$ is centered between $10^{-5}$ and $10^{-6}$, while the faulty waveforms between $10^{-1}$ and $1$, and this clear separation is being reflected in the corresponding ROC that has AUC = 0.98, and 0.96 respectively. 
The other faults show reasonable separation with an AUC values ranging from 0.83 to 0.93. 
It is important to mention that the overlapping between $normal$ and $abnormal$ KDE are because those waveforms are $normal-like$ samples and do not carry any implication that an anomaly is going to occur in the system, knowing that we are using pre-fault pulses as discussed in Section~\ref{DATA:lab}. For some faults, it may not have early enough pre-cursor/fault-indicator in the pre-fault pulse to forecasting the coming fault, thus we are not able to identify them here.
Having evaluated the multi-module results, now we compare it with the single-module approach. In Figure~\ref{fig:normal_mse}, we show normalized KDE between the two methods. 
As expected, the multi-module is learning from the multiple modules and produce lower MSE values than single-module when reconstructing $normal$ waveforms. 
The Figure shows the overall performance between the two approaches and have a side-by-side comparison between the $reconstruction~error$ values for all modules.
In Figure~\ref{fig:auc}, we show the AUC values for six fault types that have the highest number of statistics shown in Figure \ref{fig:fault} across multiple modules. 
We can see for almost all scenarios that multi-module approach has higher AUC values over the single-modules. The error bar is generated by using the probabilistic $encoder$ model that provides a probability distribution given a mean and variance parameters. By sampling from the estimated parameters at inference time, we generated multiple replicas of reconstructed $normal$ and $abnormal$ waveforms, then use the mean and SD of the replicas.
\begin{equation} \label{eq:1}
MSE = \frac{1}{n} \sum_{i=1}^n (x_{i}-\hat{x}_{i})^2
\end{equation}
\subsection{Loss Landscape}
\label{ch:loss}
In a simple ML problem, NNs are trained on feature vectors $x_i$, and corresponding labels $y_i$, and number of data samples $m$, with model parameters $\theta$, and the objective function can be defined as,
\begin{equation}
    \mathcal{L}_{\theta}=\frac{1}{m} \sum_{i=1}^m l(x_{i},y_{i},\theta)
\end{equation}

that measures how well the NN can predict $y_i$ given $x_i$ with the model weights $\theta$. The weights of NNs can be impacted by several factors, such as variable initialization, optimizer, network architecture, batch size, and other hyper-parameters. Studying the effects of various hyper-parameters is challenging because their loss values live in a high-dimensional space. Several scientific applications rely on a simple (1D line) loss curve which is computing the mean or sum of the loss value for each epoch. This produces a scalar for each iteration, and then plot the loss values as a function of epochs. While this method is beneficial to give an overview of the model performance, it only shows a small range of gradients of the parameters, and it does not show the convexity of the function, and why certain NNs architectures generalize better than others. Recently,~\citep{NEURIPS2018_a41b3bb3} devolved $filter~normalization$ technique to visualize the loss landscape of CNNs that can show how convex/non-convex an NN function is, and explain why certain NN architectures generalize well while others suffer from high generalization errors. To use this approach, one can choose a center point in the loss surface $\theta$ and choose two random direction vectors $\gamma$ and $\nu$ and plot 2D surface of the form $f(\alpha, \beta) = L(\theta + \alpha\gamma+\beta\nu)$, where $\alpha$ and $\beta$ are the grids of the loss surface and a typical value is between $-1$ and $1$. Increasing the grid values, will increase the space to compute the loss values, while shrinking the grids will zoom in and focus on the area around the minimizer of the trained model. Because the plots are sensitive to the scale of the model weights, the author suggests to rescale the random directions using the proposed $filter~normalization$ method to have the same Frobenius norm of the corresponding filter in $\theta$. This technique has been used to study the effects of different NNs architectures (e.g skip connections) and to investigate sharp versus flat minimizers and how they correlate with generalization error.
\subsubsection{Single-module vs multi-module}
\begin{figure}
\includegraphics[width=\textwidth]{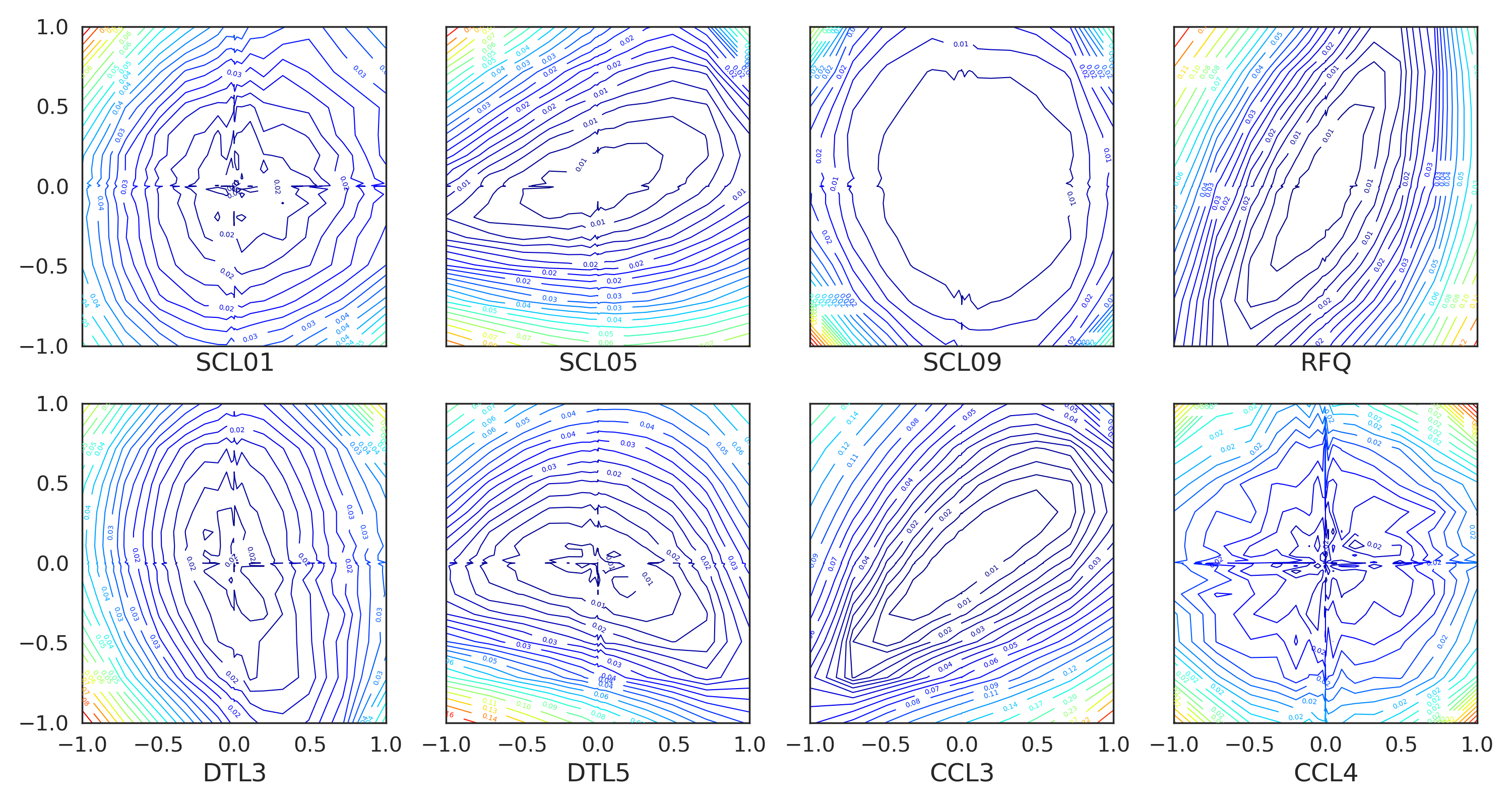}
\caption{2D contour plots of the loss surface of the single-module based VAEs using the architecture in Figure~\ref{fig:VAE}. The x- and y-axes represent the two random directions in the weight space, and the center corresponds to the model minimizer. Several loss landscapes are dominated by a region with convex contours, while others have less convexity (e.g CCL4).}
\label{fig:loss_single}
\end{figure}

\begin{figure}
\includegraphics[width=\textwidth]{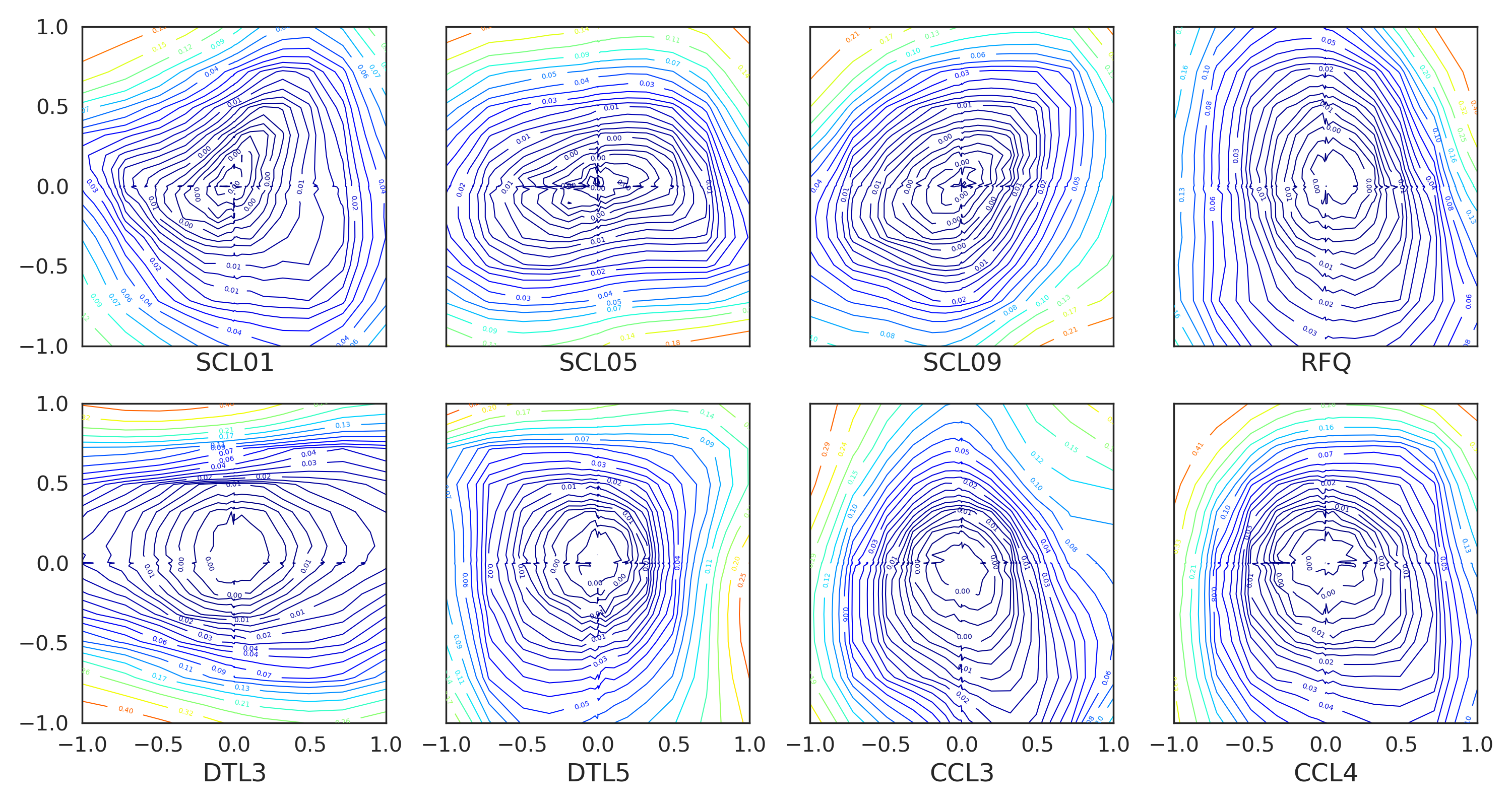}
\caption{2D contour plots of the loss surface of the multi-module CVAE based using the architecture in Figure~\ref{fig:VAE}. The x- and y-axes represent the two random directions in the weight space, and the center corresponds to the model minimizer. All examples have smooth loss surface with convex-like loss landscape.}
\label{fig:loss_multi}
\vspace{-0.3cm}
\end{figure}

In this Section, we evaluate the performance of the proposed approach multi-module CVAE and compare the model loss surface with the single-module VAE. 
To have a side-by-side comparison, for single-module we trained 15 individual VAEs and produced the corresponding loss surface, while for the multi-module we trained the model once and generated 15 loss surfaces by feeding different modules at inference time to generate the corresponding loss landscapes. 
For this exercise, we fixed the random weights initialization and used the parameters in \ref{tab:hyperparam}. 
We can see for the single-module shown in Figure~\ref{fig:loss_single} some VAE models show convex behaviour such as RFQ and CCL3, while others have chaotic loss surface such as CCL4. 
It it important to mention that there might exist an optimal model architecture for each module and this will require an extensive Neural Architecture Search (NAS) that we leave for future studies. 
For this analysis, we have instead tested the stability of the results by running different trials with different weights initialization using the same NN architecture and found that the results are consistent and produce similar loss surface. 
In Figure~\ref{fig:loss_multi}, the results obtained from the multi-module show convex-like loss surface regardless to the module we use to compute the loss landscape. 
This consistency in all modules agrees with the obtained results discussed in Section \ref{ch:res}. Evaluating each model using the loss surface was an important step to select our models in addition to analysing the anomaly detection classification accuracy. 
%
\section{Conclusion}
\label{ch:conclusion}
In this paper, we have presented multi-module based CVAE to detect HVCMs anomalies coming from different modules. The HVCMs consists of 15 modules that produce high quality neutron beams at the SNS. 
A fault in one or more of the modules can cause a major loss in operational time. 
To reduce the downtime caused by the HVCMs, our developed multi-module CVAE learns different representation of $normal$ waveforms to detect several types of anomalies and produce well-performed AD prediction ahead of time. Through our analysis, we found the multi-module CVAE model to be more sensitive to detecting anomalies than single-module VAE approach. The multi-module produces lower $reconstruction~error$ for $normal$ waveforms with an average of $10^{-4}$ as compared to  
single-module that has an average of $10^{-3}$. This increases the separation between the $reconstruction-errors$ of $normal$ and $abnormal$ samples and produces higher accuracy in detecting abnormalities. For example, for DV/DT fault, multi-module achieves AUC values ranging from 0.94 to 1.00 for all modules, while single-module VAE produces lower than 0.89 AUC as in SCL18. Similarly, for SNS PPS fault, there is more than 20\% improvement on SCL09 module, increasing from 0.73 to 1.0 AUC value. The multi-module can be trained once which eliminates the need of retraining a model for each module separately, and allow us to perform NAS and HPO efficiently. The results of the multi-module loss landscape analysis shows convex-like function for all modules, while single-module has chaotic loss surface for some cases as in CCL module. This indicates that with the given model configurations, multi-module CVAE can produce more staple results and have lower generalization error. The proposed multi-module CVAE approach can improve the HVCM reliability and allows us to schedule preventative maintenance before a catastrophic failure occurs.  
%
\section{Acknowledgement}{The authors acknowledge the help from David Brown in evaluating Operations requirements, Frank Liu, for his assistance on the Machine Learning techniques, and Sarah Cousineau for making this grant work possible. This manuscript has been authored by UT-Battelle, LLC, under contract DE-AC05-00OR22725 with the US Department of Energy (DOE). The Jefferson Science Associates (JSA) operates the Thomas Jefferson National Accelerator Facility for the U.S. Department of Energy under Contract No. DE-AC05-06OR23177. This research used resources at the Spallation Neutron Source, a DOE Office of Science User Facility operated by the Oak Ridge National Laboratory. The US government retains and the publisher, by accepting the article for publication, acknowledges that the US government retains a nonexclusive, paid-up, irrevocable, worldwide license to publish or reproduce the published form of this manuscript, or allow others to do so, for US government purposes. DOE will provide public access to these results of federally sponsored research in accordance with the DOE Public Access Plan (http://energy.gov/downloads/doe-public-access-plan).}
%
\newpage

\bibliographystyle{elsarticle-harv} 
\bibliography{main}

\begin{thebibliography}{41}
\expandafter\ifx\csname natexlab\endcsname\relax\def\natexlab#1{#1}\fi
\providecommand{\url}[1]{\texttt{#1}}
\providecommand{\href}[2]{#2}
\providecommand{\path}[1]{#1}
\providecommand{\DOIprefix}{doi:}
\providecommand{\ArXivprefix}{arXiv:}
\providecommand{\URLprefix}{URL: }
\providecommand{\Pubmedprefix}{pmid:}
\providecommand{\doi}[1]{\href{http://dx.doi.org/#1}{\path{#1}}}
\providecommand{\Pubmed}[1]{\href{pmid:#1}{\path{#1}}}
\providecommand{\bibinfo}[2]{#2}
\ifx\xfnm\relax \def\xfnm[#1]{\unskip,\space#1}\fi
\bibitem[{Abadi et~al.(2015)Abadi, Agarwal, Barham, Brevdo, Chen, Citro,
  Corrado, Davis, Dean, Devin, Ghemawat, Goodfellow, Harp, Irving, Isard, Jia,
  Jozefowicz, Kaiser, Kudlur, Levenberg, Man\'{e}, Monga, Moore, Murray, Olah,
  Schuster, Shlens, Steiner, Sutskever, Talwar, Tucker, Vanhoucke, Vasudevan,
  Vi\'{e}gas, Vinyals, Warden, Wattenberg, Wicke, Yu and
  Zheng}]{tensorflow2015-whitepaper}
\bibinfo{author}{Abadi, M.}, \bibinfo{author}{Agarwal, A.},
  \bibinfo{author}{Barham, P.}, \bibinfo{author}{Brevdo, E.},
  \bibinfo{author}{Chen, Z.}, \bibinfo{author}{Citro, C.},
  \bibinfo{author}{Corrado, G.S.}, \bibinfo{author}{Davis, A.},
  \bibinfo{author}{Dean, J.}, \bibinfo{author}{Devin, M.},
  \bibinfo{author}{Ghemawat, S.}, \bibinfo{author}{Goodfellow, I.},
  \bibinfo{author}{Harp, A.}, \bibinfo{author}{Irving, G.},
  \bibinfo{author}{Isard, M.}, \bibinfo{author}{Jia, Y.},
  \bibinfo{author}{Jozefowicz, R.}, \bibinfo{author}{Kaiser, L.},
  \bibinfo{author}{Kudlur, M.}, \bibinfo{author}{Levenberg, J.},
  \bibinfo{author}{Man\'{e}, D.}, \bibinfo{author}{Monga, R.},
  \bibinfo{author}{Moore, S.}, \bibinfo{author}{Murray, D.},
  \bibinfo{author}{Olah, C.}, \bibinfo{author}{Schuster, M.},
  \bibinfo{author}{Shlens, J.}, \bibinfo{author}{Steiner, B.},
  \bibinfo{author}{Sutskever, I.}, \bibinfo{author}{Talwar, K.},
  \bibinfo{author}{Tucker, P.}, \bibinfo{author}{Vanhoucke, V.},
  \bibinfo{author}{Vasudevan, V.}, \bibinfo{author}{Vi\'{e}gas, F.},
  \bibinfo{author}{Vinyals, O.}, \bibinfo{author}{Warden, P.},
  \bibinfo{author}{Wattenberg, M.}, \bibinfo{author}{Wicke, M.},
  \bibinfo{author}{Yu, Y.}, \bibinfo{author}{Zheng, X.}, \bibinfo{year}{2015}.
\newblock \bibinfo{title}{{TensorFlow}: Large-scale machine learning on
  heterogeneous systems}.
\newblock \URLprefix \url{https://www.tensorflow.org/}. \bibinfo{note}{software
  available from tensorflow.org}.
\bibitem[{Abdar et~al.(2021)Abdar, Pourpanah, Hussain, Rezazadegan, Liu,
  Ghavamzadeh, Fieguth, Cao, Khosravi, Acharya, Makarenkov and
  Nahavandi}]{Abdar_2021}
\bibinfo{author}{Abdar, M.}, \bibinfo{author}{Pourpanah, F.},
  \bibinfo{author}{Hussain, S.}, \bibinfo{author}{Rezazadegan, D.},
  \bibinfo{author}{Liu, L.}, \bibinfo{author}{Ghavamzadeh, M.},
  \bibinfo{author}{Fieguth, P.}, \bibinfo{author}{Cao, X.},
  \bibinfo{author}{Khosravi, A.}, \bibinfo{author}{Acharya, U.R.},
  \bibinfo{author}{Makarenkov, V.}, \bibinfo{author}{Nahavandi, S.},
  \bibinfo{year}{2021}.
\newblock \bibinfo{title}{A review of uncertainty quantification in deep
  learning: Techniques, applications and challenges}.
\newblock \bibinfo{journal}{Information Fusion} \bibinfo{volume}{76},
  \bibinfo{pages}{243--297}.
\newblock \URLprefix \url{https://doi.org/10.10162Fj.inffus.2021.05.008},
  \DOIprefix\doi{10.1016/j.inffus.2021.05.008}.
\bibitem[{Adewumi and Akinyelu(2017)}]{Adewumi2017ASO}
\bibinfo{author}{Adewumi, A.O.}, \bibinfo{author}{Akinyelu, A.A.},
  \bibinfo{year}{2017}.
\newblock \bibinfo{title}{A survey of machine-learning and nature-inspired
  based credit card fraud detection techniques}.
\newblock \bibinfo{journal}{International Journal of System Assurance
  Engineering and Management} \bibinfo{volume}{8}, \bibinfo{pages}{937--953}.
\bibitem[{Alfeo et~al.(2020)Alfeo, Cimino, Manco, Ritacco and
  Vaglini}]{ALFEO2020272}
\bibinfo{author}{Alfeo, A.L.}, \bibinfo{author}{Cimino, M.G.},
  \bibinfo{author}{Manco, G.}, \bibinfo{author}{Ritacco, E.},
  \bibinfo{author}{Vaglini, G.}, \bibinfo{year}{2020}.
\newblock \bibinfo{title}{Using an autoencoder in the design of an anomaly
  detector for smart manufacturing}.
\newblock \bibinfo{journal}{Pattern Recognition Letters} \bibinfo{volume}{136},
  \bibinfo{pages}{272--278}.
\newblock \URLprefix
  \url{https://www.sciencedirect.com/science/article/pii/S0167865520302269},
  \DOIprefix\doi{https://doi.org/10.1016/j.patrec.2020.06.008}.
\bibitem[{An and Cho(2015)}]{vae_anomaly}
\bibinfo{author}{An, J.}, \bibinfo{author}{Cho, S.}, \bibinfo{year}{2015}.
\newblock \bibinfo{title}{Variational autoencoder based anomaly detection using
  reconstruction probability}.
\newblock \bibinfo{journal}{Special Lecture on IE} \bibinfo{volume}{2}.
\bibitem[{Ball et~al.(2017)Ball, Anderson and Chan}]{Ball_2017}
\bibinfo{author}{Ball, J.E.}, \bibinfo{author}{Anderson, D.T.},
  \bibinfo{author}{Chan, C.S.}, \bibinfo{year}{2017}.
\newblock \bibinfo{title}{Comprehensive survey of deep learning in remote
  sensing: theories, tools, and challenges for the community}.
\newblock \bibinfo{journal}{Journal of Applied Remote Sensing}
  \bibinfo{volume}{11}, \bibinfo{pages}{1}.
\newblock \URLprefix \url{https://doi.org/10.11172F1.jrs.11.042609},
  \DOIprefix\doi{10.1117/1.jrs.11.042609}.
\bibitem[{Blei et~al.(2017)Blei, Kucukelbir and
  McAuliffe}]{variationalinference}
\bibinfo{author}{Blei, D.}, \bibinfo{author}{Kucukelbir, A.},
  \bibinfo{author}{McAuliffe, J.}, \bibinfo{year}{2017}.
\newblock \bibinfo{title}{Variational inference: A review for statisticians}.
\newblock \bibinfo{journal}{Journal of the American Statistical Association}
  \bibinfo{volume}{112}, \bibinfo{pages}{859--877}.
\bibitem[{Blokland et~al.(2022)Blokland, Rajput, Schram, Jeske, Ramuhalli,
  Peters, Yucesan and Zhukov}]{PhysRevAccelBeams.25.122802}
\bibinfo{author}{Blokland, W.}, \bibinfo{author}{Rajput, K.},
  \bibinfo{author}{Schram, M.}, \bibinfo{author}{Jeske, T.},
  \bibinfo{author}{Ramuhalli, P.}, \bibinfo{author}{Peters, C.},
  \bibinfo{author}{Yucesan, Y.}, \bibinfo{author}{Zhukov, A.},
  \bibinfo{year}{2022}.
\newblock \bibinfo{title}{Uncertainty aware anomaly detection to predict errant
  beam pulses in the oak ridge spallation neutron source accelerator}.
\newblock \bibinfo{journal}{Phys. Rev. Accel. Beams} \bibinfo{volume}{25},
  \bibinfo{pages}{122802}.
\newblock \URLprefix
  \url{https://link.aps.org/doi/10.1103/PhysRevAccelBeams.25.122802},
  \DOIprefix\doi{10.1103/PhysRevAccelBeams.25.122802}.
\bibitem[{Breiman(2001)}]{breiman2001random}
\bibinfo{author}{Breiman, L.}, \bibinfo{year}{2001}.
\newblock \bibinfo{title}{Random forests}.
\newblock \bibinfo{journal}{Machine Learning} \bibinfo{volume}{45},
  \bibinfo{pages}{5--32}.
\newblock \URLprefix \url{http://dx.doi.org/10.1023/A3A1010933404324},
  \DOIprefix\doi{10.1023/A:1010933404324}.
\bibitem[{Caryotakis(2004)}]{Caryotakis2004HighPK}
\bibinfo{author}{Caryotakis, G.}, \bibinfo{year}{2004}.
\newblock \bibinfo{title}{High power klystrons: Theory and practice at the
  stanford linear accelerator centerpart i}.
\bibitem[{Chalapathy and
  Chawla(2019)}]{https://doi.org/10.48550/arxiv.1901.03407}
\bibinfo{author}{Chalapathy, R.}, \bibinfo{author}{Chawla, S.},
  \bibinfo{year}{2019}.
\newblock \bibinfo{title}{Deep learning for anomaly detection: A survey}.
\newblock \URLprefix \url{https://arxiv.org/abs/1901.03407},
  \DOIprefix\doi{10.48550/ARXIV.1901.03407}.
\bibitem[{Chandola et~al.(2009)Chandola, Banerjee and
  Kumar}]{10.1145/1541880.1541882}
\bibinfo{author}{Chandola, V.}, \bibinfo{author}{Banerjee, A.},
  \bibinfo{author}{Kumar, V.}, \bibinfo{year}{2009}.
\newblock \bibinfo{title}{Anomaly detection: A survey}.
\newblock \bibinfo{journal}{ACM Comput. Surv.} \bibinfo{volume}{41}.
\newblock \URLprefix \url{https://doi.org/10.1145/1541880.1541882},
  \DOIprefix\doi{10.1145/1541880.1541882}.
\bibitem[{Chollet et~al.(2015)}]{chollet2015keras}
\bibinfo{author}{Chollet, F.}, et~al., \bibinfo{year}{2015}.
\newblock \bibinfo{title}{Keras}.
\newblock \bibinfo{howpublished}{\url{https://github.com/fchollet/keras}}.
\bibitem[{Chung et~al.(2021)Chung, Char, Guo, Schneider and
  Neiswanger}]{chung2021uncertainty}
\bibinfo{author}{Chung, Y.}, \bibinfo{author}{Char, I.}, \bibinfo{author}{Guo,
  H.}, \bibinfo{author}{Schneider, J.}, \bibinfo{author}{Neiswanger, W.},
  \bibinfo{year}{2021}.
\newblock \bibinfo{title}{Uncertainty toolbox: an open-source library for
  assessing, visualizing, and improving uncertainty quantification}.
\newblock \href{http://arxiv.org/abs/2109.10254}{{\tt arXiv:2109.10254}}.
\bibitem[{Dang et~al.(2015)Dang, Ngan and Liu}]{7251924}
\bibinfo{author}{Dang, T.T.}, \bibinfo{author}{Ngan, H.Y.},
  \bibinfo{author}{Liu, W.}, \bibinfo{year}{2015}.
\newblock \bibinfo{title}{Distance-based k-nearest neighbors outlier detection
  method in large-scale traffic data}, in: \bibinfo{booktitle}{2015 IEEE
  International Conference on Digital Signal Processing (DSP)}, pp.
  \bibinfo{pages}{507--510}.
\newblock \DOIprefix\doi{10.1109/ICDSP.2015.7251924}.
\bibitem[{Edelen et~al.(2018)Edelen, Mayes, Bowring, Ratner, Adelmann,
  Ischebeck, Snuverink, Agapov, Kammering, Edelen
  et~al.}]{edelen2018opportunities}
\bibinfo{author}{Edelen, A.}, \bibinfo{author}{Mayes, C.},
  \bibinfo{author}{Bowring, D.}, \bibinfo{author}{Ratner, D.},
  \bibinfo{author}{Adelmann, A.}, \bibinfo{author}{Ischebeck, R.},
  \bibinfo{author}{Snuverink, J.}, \bibinfo{author}{Agapov, I.},
  \bibinfo{author}{Kammering, R.}, \bibinfo{author}{Edelen, J.}, et~al.,
  \bibinfo{year}{2018}.
\newblock \bibinfo{title}{Opportunities in machine learning for particle
  accelerators}.
\newblock \bibinfo{journal}{arXiv preprint arXiv:1811.03172} .
\bibitem[{Edelen and Cook(2021)}]{edelen2021anomaly}
\bibinfo{author}{Edelen, J.P.}, \bibinfo{author}{Cook, N.M.},
  \bibinfo{year}{2021}.
\newblock \bibinfo{title}{Anomaly detection in particle accelerators using
  autoencoders}.
\newblock \bibinfo{journal}{arXiv preprint arXiv:2112.07793} .
\bibitem[{Hinton and Zemel(1993)}]{10.5555/2987189.2987190}
\bibinfo{author}{Hinton, G.E.}, \bibinfo{author}{Zemel, R.S.},
  \bibinfo{year}{1993}.
\newblock \bibinfo{title}{Autoencoders, minimum description length and
  helmholtz free energy}, in: \bibinfo{booktitle}{Proceedings of the 6th
  International Conference on Neural Information Processing Systems},
  \bibinfo{publisher}{Morgan Kaufmann Publishers Inc.}, \bibinfo{address}{San
  Francisco, CA, USA}. p. \bibinfo{pages}{3–10}.
\bibitem[{John(1995)}]{John_outliers}
\bibinfo{author}{John, G.H.}, \bibinfo{year}{1995}.
\newblock \bibinfo{title}{Robust decision trees: Removing outliers from
  databases}, in: \bibinfo{booktitle}{Proceedings of the First International
  Conference on Knowledge Discovery and Data Mining}, \bibinfo{publisher}{AAAI
  Press}. p. \bibinfo{pages}{174–179}.
\bibitem[{Kieu et~al.(2019)Kieu, Yang, Guo and Jensen}]{ijcai2019-378}
\bibinfo{author}{Kieu, T.}, \bibinfo{author}{Yang, B.}, \bibinfo{author}{Guo,
  C.}, \bibinfo{author}{Jensen, C.S.}, \bibinfo{year}{2019}.
\newblock \bibinfo{title}{Outlier detection for time series with recurrent
  autoencoder ensembles}, in: \bibinfo{booktitle}{Proceedings of the
  Twenty-Eighth International Joint Conference on Artificial Intelligence,
  {IJCAI-19}}, \bibinfo{publisher}{International Joint Conferences on
  Artificial Intelligence Organization}. pp. \bibinfo{pages}{2725--2732}.
\newblock \URLprefix \url{https://doi.org/10.24963/ijcai.2019/378},
  \DOIprefix\doi{10.24963/ijcai.2019/378}.
\bibitem[{Kingma and Welling(2013)}]{https://doi.org/10.48550/arxiv.1312.6114}
\bibinfo{author}{Kingma, D.P.}, \bibinfo{author}{Welling, M.},
  \bibinfo{year}{2013}.
\newblock \bibinfo{title}{Auto-encoding variational bayes}.
\newblock \URLprefix \url{https://arxiv.org/abs/1312.6114},
  \DOIprefix\doi{10.48550/ARXIV.1312.6114}.
\bibitem[{Koch et~al.(2015)Koch, Zemel and Salakhutdinov}]{Koch2015SiameseNN}
\bibinfo{author}{Koch, G.}, \bibinfo{author}{Zemel, R.},
  \bibinfo{author}{Salakhutdinov, R.}, \bibinfo{year}{2015}.
\newblock \bibinfo{title}{Siamese neural networks for one-shot image
  recognition}.
\bibitem[{Krizhevsky et~al.(2012)Krizhevsky, Sutskever and
  Hinton}]{NIPS2012_c399862d}
\bibinfo{author}{Krizhevsky, A.}, \bibinfo{author}{Sutskever, I.},
  \bibinfo{author}{Hinton, G.E.}, \bibinfo{year}{2012}.
\newblock \bibinfo{title}{Imagenet classification with deep convolutional
  neural networks}, in: \bibinfo{editor}{Pereira, F.}, \bibinfo{editor}{Burges,
  C.}, \bibinfo{editor}{Bottou, L.}, \bibinfo{editor}{Weinberger, K.} (Eds.),
  \bibinfo{booktitle}{Advances in Neural Information Processing Systems},
  \bibinfo{publisher}{Curran Associates, Inc.}
\newblock \URLprefix
  \url{https://proceedings.neurips.cc/paper/2012/file/c399862d3b9d6b76c8436e924a68c45b-Paper.pdf}.
\bibitem[{Kullback and Leibler(1951)}]{kullback1951information}
\bibinfo{author}{Kullback, S.}, \bibinfo{author}{Leibler, R.A.},
  \bibinfo{year}{1951}.
\newblock \bibinfo{title}{On information and sufficiency}.
\newblock \bibinfo{journal}{The annals of mathematical statistics}
  \bibinfo{volume}{22}, \bibinfo{pages}{79--86}.
\bibitem[{LeCun and Bengio(1998)}]{10.5555/303568.303704}
\bibinfo{author}{LeCun, Y.}, \bibinfo{author}{Bengio, Y.},
  \bibinfo{year}{1998}.
\newblock \bibinfo{title}{Convolutional Networks for Images, Speech, and Time
  Series}. \bibinfo{publisher}{MIT Press}, \bibinfo{address}{Cambridge, MA,
  USA}.
\newblock p. \bibinfo{pages}{255–258}.
\bibitem[{Li et~al.(2018)Li, Xu, Taylor, Studer and
  Goldstein}]{NEURIPS2018_a41b3bb3}
\bibinfo{author}{Li, H.}, \bibinfo{author}{Xu, Z.}, \bibinfo{author}{Taylor,
  G.}, \bibinfo{author}{Studer, C.}, \bibinfo{author}{Goldstein, T.},
  \bibinfo{year}{2018}.
\newblock \bibinfo{title}{Visualizing the loss landscape of neural nets}, in:
  \bibinfo{editor}{Bengio, S.}, \bibinfo{editor}{Wallach, H.},
  \bibinfo{editor}{Larochelle, H.}, \bibinfo{editor}{Grauman, K.},
  \bibinfo{editor}{Cesa-Bianchi, N.}, \bibinfo{editor}{Garnett, R.} (Eds.),
  \bibinfo{booktitle}{Advances in Neural Information Processing Systems},
  \bibinfo{publisher}{Curran Associates, Inc.}
\newblock \URLprefix
  \url{https://proceedings.neurips.cc/paper/2018/file/a41b3bb3e6b050b6c9067c67f663b915-Paper.pdf}.
\bibitem[{Litjens et~al.(2017)Litjens, Kooi, Bejnordi, Setio, Ciompi,
  Ghafoorian, {van der Laak}, {van Ginneken} and Sánchez}]{LITJENS201760}
\bibinfo{author}{Litjens, G.}, \bibinfo{author}{Kooi, T.},
  \bibinfo{author}{Bejnordi, B.E.}, \bibinfo{author}{Setio, A.A.A.},
  \bibinfo{author}{Ciompi, F.}, \bibinfo{author}{Ghafoorian, M.},
  \bibinfo{author}{{van der Laak}, J.A.}, \bibinfo{author}{{van Ginneken}, B.},
  \bibinfo{author}{Sánchez, C.I.}, \bibinfo{year}{2017}.
\newblock \bibinfo{title}{A survey on deep learning in medical image analysis}.
\newblock \bibinfo{journal}{Medical Image Analysis} \bibinfo{volume}{42},
  \bibinfo{pages}{60--88}.
\newblock \URLprefix
  \url{https://www.sciencedirect.com/science/article/pii/S1361841517301135},
  \DOIprefix\doi{https://doi.org/10.1016/j.media.2017.07.005}.
\bibitem[{Lu et~al.(2017)Lu, Cheng, Xiao, Chang, Huang, Liang and
  Huang}]{10.1109/TIP.2017.2713048}
\bibinfo{author}{Lu, W.}, \bibinfo{author}{Cheng, Y.}, \bibinfo{author}{Xiao,
  C.}, \bibinfo{author}{Chang, S.}, \bibinfo{author}{Huang, S.},
  \bibinfo{author}{Liang, B.}, \bibinfo{author}{Huang, T.},
  \bibinfo{year}{2017}.
\newblock \bibinfo{title}{Unsupervised sequential outlier detection with deep
  architectures}.
\newblock \bibinfo{journal}{Trans. Img. Proc.} \bibinfo{volume}{26},
  \bibinfo{pages}{4321–4330}.
\newblock \URLprefix \url{https://doi.org/10.1109/TIP.2017.2713048},
  \DOIprefix\doi{10.1109/TIP.2017.2713048}.
\bibitem[{Marcato et~al.(2021)Marcato, Arena, Bortolato, Gelain, Martinelli,
  Munaron, Roetta, Savarese and Susto}]{marcato2021machine}
\bibinfo{author}{Marcato, D.}, \bibinfo{author}{Arena, G.},
  \bibinfo{author}{Bortolato, D.}, \bibinfo{author}{Gelain, F.},
  \bibinfo{author}{Martinelli, V.}, \bibinfo{author}{Munaron, E.},
  \bibinfo{author}{Roetta, M.}, \bibinfo{author}{Savarese, G.},
  \bibinfo{author}{Susto, G.A.}, \bibinfo{year}{2021}.
\newblock \bibinfo{title}{Machine learning-based anomaly detection for particle
  accelerators}, in: \bibinfo{booktitle}{2021 IEEE Conference on Control
  Technology and Applications (CCTA)}, \bibinfo{organization}{IEEE}. pp.
  \bibinfo{pages}{240--246}.
\bibitem[{Mohammadi et~al.(2017)Mohammadi, Al-Fuqaha, Sorour and
  Guizani}]{https://doi.org/10.48550/arxiv.1712.04301}
\bibinfo{author}{Mohammadi, M.}, \bibinfo{author}{Al-Fuqaha, A.},
  \bibinfo{author}{Sorour, S.}, \bibinfo{author}{Guizani, M.},
  \bibinfo{year}{2017}.
\newblock \bibinfo{title}{Deep learning for iot big data and streaming
  analytics: A survey}.
\newblock \URLprefix \url{https://arxiv.org/abs/1712.04301},
  \DOIprefix\doi{10.48550/ARXIV.1712.04301}.
\bibitem[{Pappas et~al.(2021)Pappas, Lu, Schram and Vrabie}]{pappas2021machine}
\bibinfo{author}{Pappas, G.}, \bibinfo{author}{Lu, D.},
  \bibinfo{author}{Schram, M.}, \bibinfo{author}{Vrabie, D.},
  \bibinfo{year}{2021}.
\newblock \bibinfo{title}{Machine learning for improved availability of the sns
  klystron high voltage converter modulators}, in: \bibinfo{booktitle}{Proc.
  IPAC'21}, \bibinfo{publisher}{JACoW Publishing, Geneva, Switzerland}. pp.
  \bibinfo{pages}{4303--4306}.
\newblock \DOIprefix\doi{10.18429/JACoW-IPAC2021-THPAB252}.
\bibitem[{Potter(2006)}]{Potter2006MethodsFP}
\bibinfo{author}{Potter, K.C.}, \bibinfo{year}{2006}.
\newblock \bibinfo{title}{Methods for presenting statistical information: The
  box plot}, in: \bibinfo{booktitle}{Visualization of Large and Unstructured
  Data Sets}.
\bibitem[{Radaideh et~al.(2022a)Radaideh, Pappas and
  Cousineau}]{radaideh2022real}
\bibinfo{author}{Radaideh, M.I.}, \bibinfo{author}{Pappas, C.},
  \bibinfo{author}{Cousineau, S.}, \bibinfo{year}{2022}a.
\newblock \bibinfo{title}{Real electronic signal data from particle accelerator
  power systems for machine learning anomaly detection}.
\newblock \bibinfo{journal}{Data in Brief} \bibinfo{volume}{43},
  \bibinfo{pages}{108473}.
\bibitem[{Radaideh et~al.(2022b)Radaideh, Pappas, Walden, Lu, Vidyaratne,
  Britton, Rajput, Schram and Cousineau}]{radaideh2022time}
\bibinfo{author}{Radaideh, M.I.}, \bibinfo{author}{Pappas, C.},
  \bibinfo{author}{Walden, J.}, \bibinfo{author}{Lu, D.},
  \bibinfo{author}{Vidyaratne, L.}, \bibinfo{author}{Britton, T.},
  \bibinfo{author}{Rajput, K.}, \bibinfo{author}{Schram, M.},
  \bibinfo{author}{Cousineau, S.}, \bibinfo{year}{2022}b.
\newblock \bibinfo{title}{Time series anomaly detection in power electronics
  signals with recurrent and convlstm autoencoders}.
\newblock \bibinfo{journal}{Digital Signal Processing} \bibinfo{volume}{130},
  \bibinfo{pages}{103704}.
\bibitem[{Radaideh et~al.(2022c)Radaideh, Tran, Lin, Jiang, Winder, Gorti,
  Zhang, Mach and Cousineau}]{radaideh2022model}
\bibinfo{author}{Radaideh, M.I.}, \bibinfo{author}{Tran, H.},
  \bibinfo{author}{Lin, L.}, \bibinfo{author}{Jiang, H.},
  \bibinfo{author}{Winder, D.}, \bibinfo{author}{Gorti, S.},
  \bibinfo{author}{Zhang, G.}, \bibinfo{author}{Mach, J.},
  \bibinfo{author}{Cousineau, S.}, \bibinfo{year}{2022}c.
\newblock \bibinfo{title}{Model calibration of the liquid mercury spallation
  target using evolutionary neural networks and sparse polynomial expansions}.
\newblock \bibinfo{journal}{Nuclear Instruments and Methods in Physics Research
  Section B: Beam Interactions with Materials and Atoms} \bibinfo{volume}{525},
  \bibinfo{pages}{41--54}.
\bibitem[{Reass et~al.(2003)Reass, Apgar, Baca, Borovina, Bradle, Doss,
  Gonzales, Gribble, Hardek, Lynch, Rees, Tallerico, Trujillo, Anderson,
  Heidenreich, Hicks and Leontiev}]{1288975}
\bibinfo{author}{Reass, W.}, \bibinfo{author}{Apgar, S.},
  \bibinfo{author}{Baca, D.}, \bibinfo{author}{Borovina, D.},
  \bibinfo{author}{Bradle, J.}, \bibinfo{author}{Doss, J.},
  \bibinfo{author}{Gonzales, J.}, \bibinfo{author}{Gribble, R.},
  \bibinfo{author}{Hardek, T.}, \bibinfo{author}{Lynch, M.},
  \bibinfo{author}{Rees, D.}, \bibinfo{author}{Tallerico, P.},
  \bibinfo{author}{Trujillo, P.}, \bibinfo{author}{Anderson, D.},
  \bibinfo{author}{Heidenreich, D.}, \bibinfo{author}{Hicks, J.},
  \bibinfo{author}{Leontiev, V.}, \bibinfo{year}{2003}.
\newblock \bibinfo{title}{Design, status, and first operations of the
  spallation neutron source polyphase resonant converter modulator system}, in:
  \bibinfo{booktitle}{Proceedings of the 2003 Particle Accelerator Conference},
  pp. \bibinfo{pages}{553--557 Vol.1}.
\newblock \DOIprefix\doi{10.1109/PAC.2003.1288975}.
\bibitem[{Rescic et~al.(2020)Rescic, Seviour and
  Blokland}]{rescic2020predicting}
\bibinfo{author}{Rescic, M.}, \bibinfo{author}{Seviour, R.},
  \bibinfo{author}{Blokland, W.}, \bibinfo{year}{2020}.
\newblock \bibinfo{title}{Predicting particle accelerator failures using binary
  classifiers}.
\newblock \bibinfo{journal}{Nuclear Instruments and Methods in Physics Research
  Section A: Accelerators, Spectrometers, Detectors and Associated Equipment}
  \bibinfo{volume}{955}, \bibinfo{pages}{163240}.
\bibitem[{Re{\v{s}}{\v{c}}i{\v{c}} et~al.(2022)Re{\v{s}}{\v{c}}i{\v{c}},
  Seviour and Blokland}]{revsvcivc2022improvements}
\bibinfo{author}{Re{\v{s}}{\v{c}}i{\v{c}}, M.}, \bibinfo{author}{Seviour, R.},
  \bibinfo{author}{Blokland, W.}, \bibinfo{year}{2022}.
\newblock \bibinfo{title}{Improvements of pre-emptive identification of
  particle accelerator failures using binary classifiers and dimensionality
  reduction}.
\newblock \bibinfo{journal}{Nuclear Instruments and Methods in Physics Research
  Section A: Accelerators, Spectrometers, Detectors and Associated Equipment}
  \bibinfo{volume}{1025}, \bibinfo{pages}{166064}.
\bibitem[{Sohn et~al.(2015)Sohn, Lee and Yan}]{NIPS2015_8d55a249}
\bibinfo{author}{Sohn, K.}, \bibinfo{author}{Lee, H.}, \bibinfo{author}{Yan,
  X.}, \bibinfo{year}{2015}.
\newblock \bibinfo{title}{Learning structured output representation using deep
  conditional generative models}, in: \bibinfo{editor}{Cortes, C.},
  \bibinfo{editor}{Lawrence, N.}, \bibinfo{editor}{Lee, D.},
  \bibinfo{editor}{Sugiyama, M.}, \bibinfo{editor}{Garnett, R.} (Eds.),
  \bibinfo{booktitle}{Advances in Neural Information Processing Systems},
  \bibinfo{publisher}{Curran Associates, Inc.}
\newblock \URLprefix
  \url{https://proceedings.neurips.cc/paper/2015/file/8d55a249e6baa5c06772297520da2051-Paper.pdf}.
\bibitem[{White(2002)}]{SNS_cite}
\bibinfo{author}{White, M.}, \bibinfo{year}{2002}.
\newblock \bibinfo{title}{The spallation neutron source (sns)}, in:
  \bibinfo{booktitle}{Proceedings of LINAC 2002, Gyeongju, Korea}.
\bibitem[{Xu et~al.(2015)Xu, Ricci, Yan, Song and
  Sebe}]{https://doi.org/10.48550/arxiv.1510.01553}
\bibinfo{author}{Xu, D.}, \bibinfo{author}{Ricci, E.}, \bibinfo{author}{Yan,
  Y.}, \bibinfo{author}{Song, J.}, \bibinfo{author}{Sebe, N.},
  \bibinfo{year}{2015}.
\newblock \bibinfo{title}{Learning deep representations of appearance and
  motion for anomalous event detection}.
\newblock \URLprefix \url{https://arxiv.org/abs/1510.01553},
  \DOIprefix\doi{10.48550/ARXIV.1510.01553}.

\end{thebibliography}
\newpage
\appendix
\section{Deep vs Shallow Neural Networks}
\begin{figure}
\includegraphics[width=\textwidth]{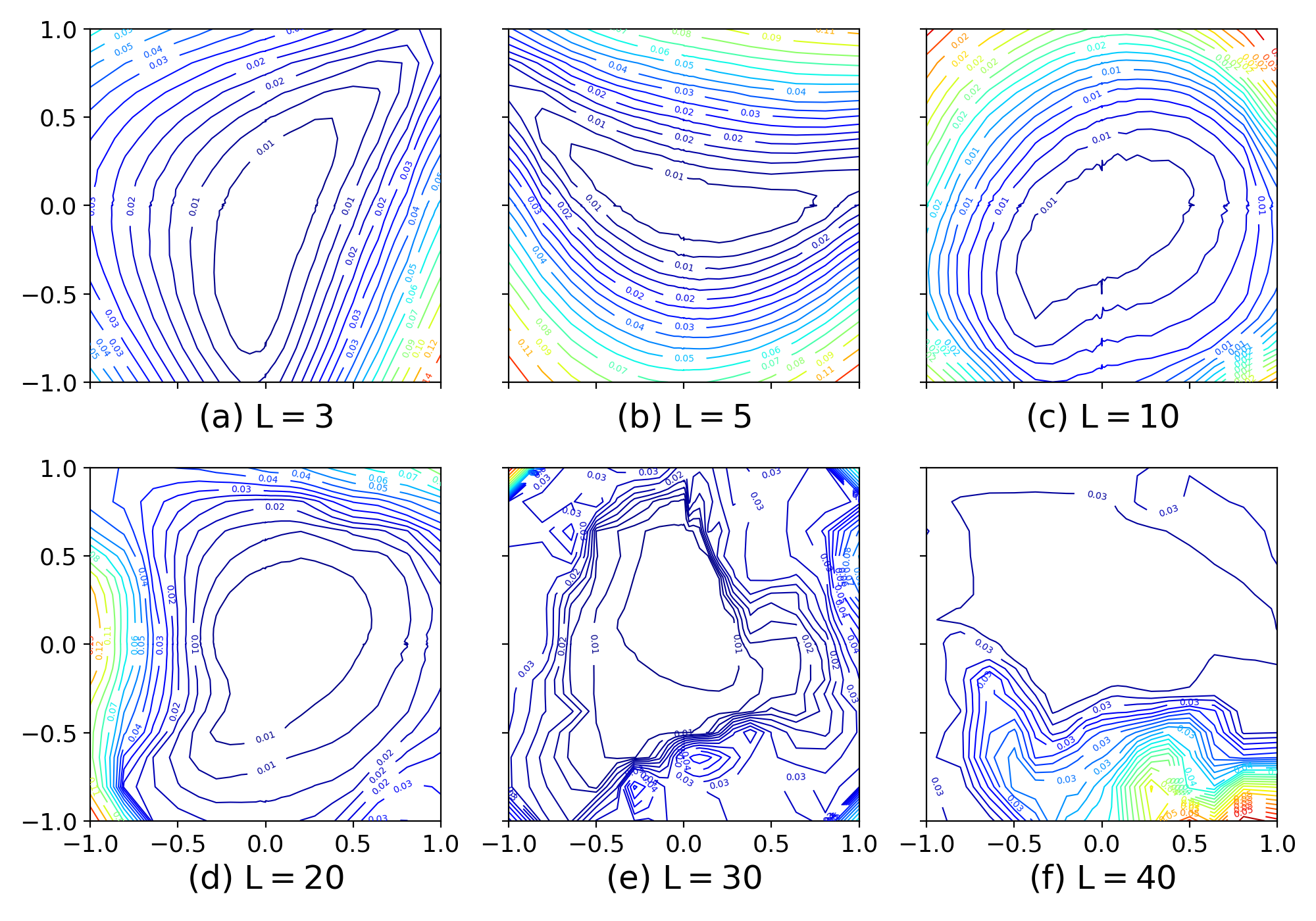}
\caption{2D visualization of the loss surface of the multi-module model trained using different number of Conv1D layers, where L is the number of layers in the $encoder$ and $decoder$. The transits from smooth to very chaotic loss surface.}
\label{fig:deepvsshallow}
\end{figure}

This section is intended to provide a visual understanding of the effects of different number of layers used in the model architecture using filter normalization technique~\citep{NEURIPS2018_a41b3bb3}.
The model was trained six times using 3, 5, 10, 20, 30, and 40 CNN layers in the $encoder$ and the $decoder$. All the other parameters are held constants to see only the effect of deep vs shallow $encoder$ and $decoder$. The hyper-parameter choice is adopted from the model discussed in \ref{SINGLE_VAE}. As we can see in Figure \ref{fig:deepvsshallow}, there is a transition from nearly convex to being highly chaotic loss surface associated with increasing the number of layers. While this behaviour is expected because the model starts to suffer from several problems (e.g. vanishing gradient descents) with very deep NN layers, it can help to select the appropriate number of layers that have convex-like loss surface. 
\section{Uncertainty quantification}
\label{ch:uq}
\begin{figure}
\includegraphics[width=\textwidth, height = 6cm]{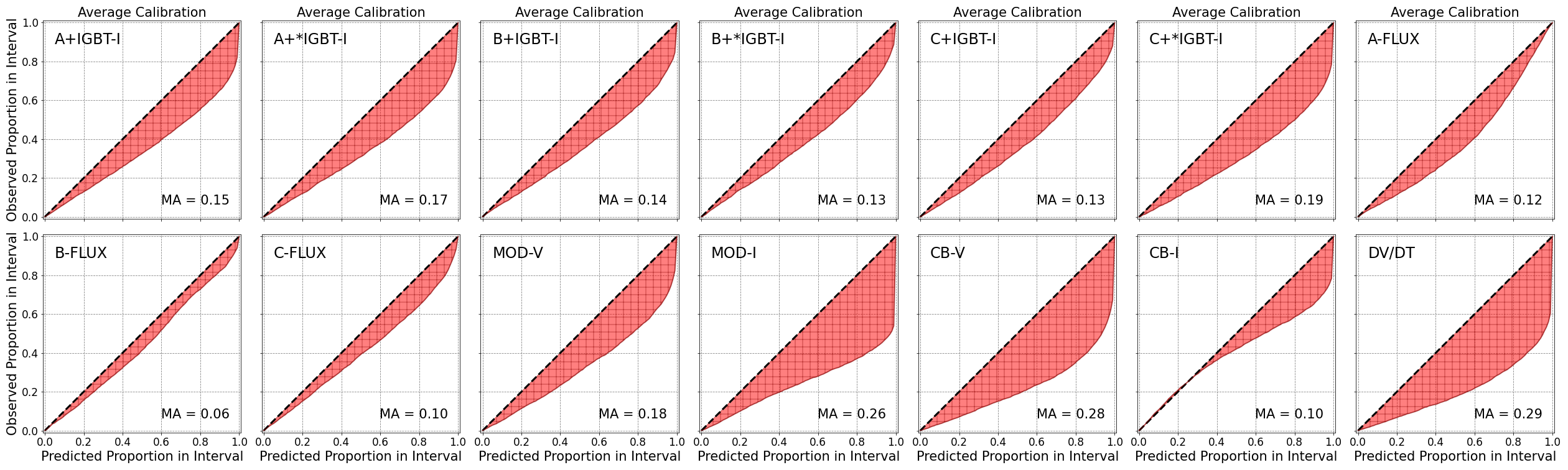}
\caption{Average Miscalibration Area (MA) for each waveform using random examples from SCL01 module. The MA ranges from 6\% to 29\%.}
\label{fig:scl18_ma}
\end{figure}

\begin{figure}
\includegraphics[width=\textwidth,  height = 6cm]{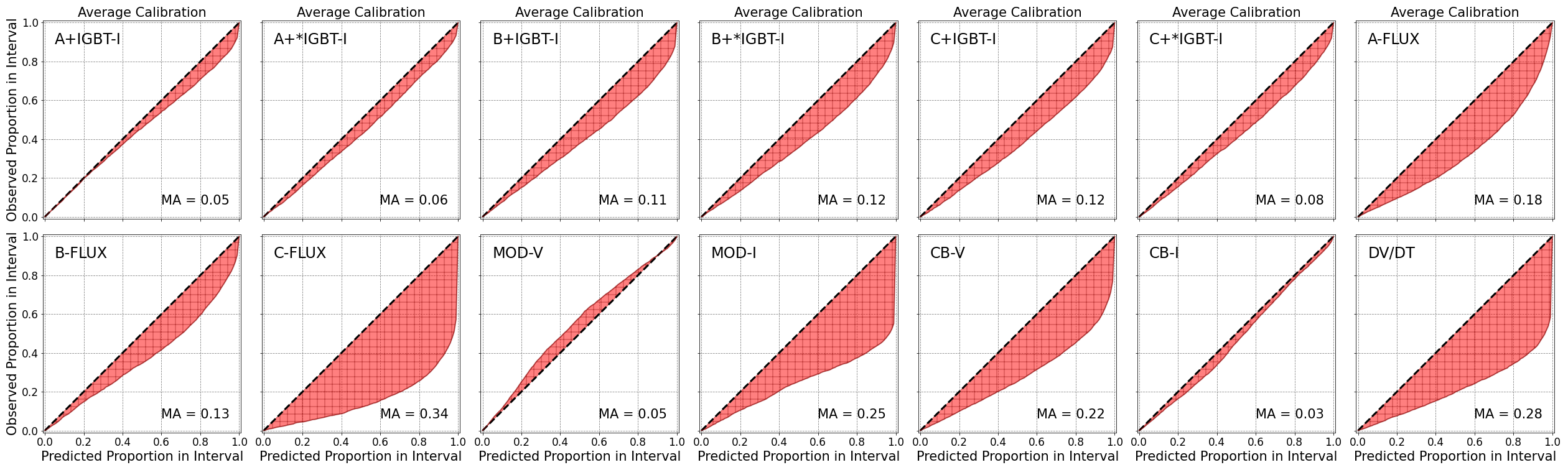}
\caption{As in Figure~\ref{fig:scl18_ma}, but using CCL4 module.}
\label{fig:ccl4_ma}
\end{figure}

There are several Uncertainty Quantification (UQ) methods used in the ML community to quantify the model and data uncertainty. 
A comprehensive survey of UQ methods in ML can be found in~\citep{Abdar_2021}. 
In this work, we generated the uncertainty for the predicted results by sampling the latent $z$ given a mean and variance parameters generated from the probabilistic $encoder$ model. Given the generated replicas of reconstructed $normal$ and $abnormal$ waveforms, we use the mean and SD to estimate the uncertainty using an uncertainty toolbox \citep{chung2021uncertainty}. The toolbox provides a miscalibration area (MA) by plotting the observed proportion verses prediction proportion of outputs falling into a range of intervals, and given a number of bins. Figure~\ref{fig:scl18_ma} and \ref{fig:ccl4_ma} show the Average Miscalibration Area (MA) from 10 ranom examples selected from SCL01 and CCL4 modules respectively. The results show that some of the waveforms, such as the IGBTs have small MA, while MOD-I and CB-V have larger MA for both modules. It is important to mention that the results are generated using few random examples and they might not represent all other examples or modules. For our future work, we plan to investigate model calibration and have an extensive study to quantify the uncertinaty of the predicted results.






\end{document}